\documentclass[final]{cvpr}
\pdfoutput=1

\usepackage[english]{babel}
\usepackage[utf8]{inputenc}
\usepackage{algorithm}
\usepackage[noend]{algpseudocode}
\usepackage{times}
\usepackage{epsfig}
\usepackage{graphicx}
\usepackage{amsmath}
\usepackage{multirow}
\usepackage{booktabs}
\usepackage{subcaption}
\DeclareMathOperator*{\argmin}{argmin} 

\usepackage{amssymb}
\usepackage{booktabs}

\usepackage[pagebackref=true,breaklinks=true,colorlinks,bookmarks=false]{hyperref}

\def\cvprPaperID{****} 
\def\confYear{CVPR 2021}

\newcommand{\s}[1]{}
\newcommand{\burak}[1]{}
\newcommand{\ayush}[1]{}
\newcommand{\evan}[1]{}
\newcommand{\shuvam}[1]{}
\begin{document}

\title{Efficient Conditional Pre-training for Transfer Learning}

\author{
Shuvam Chakraborty\\
Department of Computer Science\\
Stanford University\\
{\tt\small shuvamc@stanford.edu}
\and
Burak Uzkent\\
Department of Computer Science\\
Stanford University\\
{\tt\small buzkent@cs.stanford.edu}
\and
Kumar Ayush \\
Department of Computer Science\\
Stanford University\\
{\tt\small kayush@stanford.edu}
\and
Kumar Tanmay\\
Department of Computer Science\\
IIT Kharagpur\\
{\tt\small kr.tanmay147@iitkgp.ac.in}
\and
Evan Sheehan\\
Department of Computer Science\\
Stanford University\\
{\tt\small esheehan@stanford.edu}
\and
Stefano Ermon\\
Department of Computer Science\\
Stanford University\\
{\tt\small ermon@cs.stanford.edu}
}

\maketitle

\begin{abstract}
    Almost all the state-of-the-art neural networks for computer vision tasks are trained by (1) pre-training on a large-scale dataset and (2) finetuning on the target dataset. 
    This strategy helps reduce dependence on the target dataset and improves convergence rate and generalization on the target task. 
    Although pre-training on large-scale datasets is very useful for new methods or models, its foremost disadvantage is high training cost. To address this, we propose efficient filtering methods to select relevant subsets from the pre-training dataset. Additionally, we discover that lowering image resolutions in the pre-training step offers a great trade-off between cost and performance. We validate our techniques by pre-training on ImageNet in both the unsupervised and supervised settings and finetuning on a diverse collection of target datasets and tasks. Our proposed methods drastically reduce pre-training cost and provide strong performance boosts. Finally, we improve the current standard of ImageNet pre-training by $1\text{-}3\%$ by tuning available models on our subsets and pre-training on a dataset filtered from a larger scale dataset.
\end{abstract}

\vspace{-1em}
\section{Introduction}
Recent success of modern computer vision methods relies heavily on large-scale labelled datasets, which are often costly to collect~\cite{mahajan2018exploring,he2019momentum,chen2020improved}. Alternatives to large-scale labelled data include pre-training a network on the publicly available ImageNet dataset with labels~\cite{deng2009imagenet} and performing transfer learning on target tasks~\cite{huh2016makes,xie2015transfer,shin2016deep,hendrycks2019using,kornblith2019better}. On the other hand, unsupervised learning has received tremendous attention recently with the availability of extremely large-scale data with no labels, as such data is costly to obtain ~\cite{mahajan2018exploring,he2019momentum,he2020momentum,grill2020bootstrap,caron2020unsupervised,chen2020simple, chen2020improved}. 

The explosion of data quantity and improvement of unsupervised learning with contrastive learning portends that the standard approach in future tasks will be to (1) learn weights a on a very large-scale dataset with unsupervised learning and (2) fine-tune them on a small-scale target dataset. A major problem with this approach is the large amount of computational resources required to train a network on a very large scale dataset~\cite{mahajan2018exploring}. For example, a recent contrastive learning method, MoCo-v2~\cite{he2020momentum,he2019momentum}, uses 8 GPUs to train on ImageNet for 53 hours, which can cost thousands of dollars. Extrapolating, this forebodes pre-training costs on the order of millions of dollars on larger-scale datasets. Those without access to such computation power will require selecting relevant subsets of those datasets specific to their task. 

Cognizant of these pressing issues, we propose novel methods to efficiently filter a user defined number of pre-training images conditioned on a target dataset. 
We also find that the use of low resolution images during pre-training provides a great cost to performance trade-off. Our approach consistently outperforms other methods
by $2\text{-}9\%$ and are both flexible, translating to both supervised and unsupervised settings, and adaptable, translating to a wide range of target tasks including image recognition, object detection and semantic segmentation. Our methods perform especially well in the more relevant unsupervised setting where pre-training on a $12\%$ subset of data can achieve within $1\text{-}4\%$ of full pre-training when considering target task performance. Next, we use our methods to improve standard ImageNet (1.28M images) pre-training. In this direction, we construct a large scale dataset (6.7M images) from multiple datasets and filter 1.28M images conditioned on a target task. Our results show that we improve standard ImageNet pre-training by $1\text{-}3\%$ on downstream tasks. Thus, when needing to pre-train from scratch on large scale data for a specific application, our methods can replace the standard ImageNet pre-training 
with conditional pre-training.

\vspace{-0.5em}
\section{Related Work}
\noindent \textbf{Active Learning} Active Learning fits a function by selectively querying labels for samples where the function is currently uncertain. In a basic greedy setup, the samples with the highest entropies are chosen for annotation \cite{wang2016cost,gal2017deep,beluch2018power,sener2017active}. 
Active learning typically assumes similar data distributions for candidate samples, whereas our data distributions can potentially have large shifts. Furthermore, active learning, due to its iterative nature, can be quite costly, hard to tune, and can require prior distributions \cite{park2012bayesian}.

\noindent \textbf{Unconditional Transfer Learning} 
Pre-training networks on ImageNet has been shown to be a very effective way of initializing weights for a target task with small sample size~\cite{huh2016makes,xie2015transfer,shin2016deep,hendrycks2019using,kornblith2019better,uzkent2019ijcai,uzkent2018tracking,sheehan2018learning}. However, all these studies use unconditional pre-training as they employ the weights pre-trained on the full source dataset, which can be computationally infeasible for future large scale datasets.

\noindent \textbf{Conditional Transfer Learning} \cite{Yan_2020_CVPR,cui2018large,ngiam2018domain}, on the other hand, filter the pre-training dataset conditioned on target tasks. ~\cite{cui2018large, ge2017borrowing} use greedy class-specific clustering based and learn image representations with an encoder trained on the massive JFT-300M dataset~\cite{hinton2015distilling}, which dramatically increases cost.~\cite{Yan_2020_CVPR} trains a number of expert models on many subsets of the pre-training dataset and uses their performance to weight source images, however this method is naturally quite computationally expensive. Many of these methods also require labelled pre-training data and are not well suited for target tasks such as object detection and semantic segmentation. Our methods differ from these works as we take into account pre-training dataset filtering efficiency, adaptability to different target tasks and settings, and target task performance.

\vspace{-0.7em}
\section{Problem Definition and Setup}
We assume a target task dataset represented as $\mathcal{D}_{t}=(\mathcal{X}_{t},\mathcal{Y}_{t})$ where $\mathcal{X}_{t}=\{x_{t}^{1}, x_{t}^{2}, \dots, x_{t}^{M}\}$ represents a set of M images with their ground truth labels $\mathcal{Y}_t$. Our goal is to train a function $f_{t}$ parameterized by $\theta_{t}$ on the dataset $\mathcal{D}_{t}$ to learn $f_{t}:x_{t}^{i} \mapsto y_{t}^{i}$. To transfer learn, we first pre-train $\theta_{t}$ on a large-scale source dataset $\mathcal{D}_{s}$ and fine-tune $\theta_{t}$ on $\mathcal{D}_{t}$. This strategy reduces the amount of labeled samples needed in $\mathcal{D}_{t}$ and boosts the accuracy in comparison to the randomly initialized weights~\cite{mahajan2018exploring,uzkent2019learning}. For the pre-training dataset, we can have either labelled or unlabelled setups: (1) $\mathcal{D}_{s}=(\mathcal{X}_{s}, \mathcal{Y}_{s})$ and (2) $\mathcal{D}_{s}=(\mathcal{X}_{s})$ where $\mathcal{X}_{s}=\{x_{s}^{1}, x_{s}^{2}, \dots, x_{s}^{N}\}$. However, it is tough to label vast amounts of publicly available images, and with the increasing popularity of unsupervised learning methods\cite{chen2020improved, chen2020mocov2, chen2020simple, he2019momentum, he2020momentum}, it is easy to see that unsupervised pre-training on very large $\mathcal{D}_{s}$ with no ground-truth labels will be the standard and preferred practice in the future. 

A major problem with learning $\theta_{t}$ on a very large-scale dataset $\mathcal{D}_{s}$ is the computational cost, and using the whole dataset may be impossible for most. One way to reduce costs is to filter out images deemed less relevant for $\mathcal{D}_{t}$ to create a dataset $\mathcal{D}_{s}^{'} \in \mathcal{D}_{s}$ where $\mathcal{X}_{s}=\{x_{s}^{1}, x_{s}^{2}, \dots, x_{s}^{N^{'}}\}$ represents a filtered version of $\mathcal{D}_{s}$ with $N^{'}\ll N$. Our approach conditions the filtering step on the target dataset $\mathcal{D}_{t}$.
In this study, we propose flexible and adaptable methods to perform \emph{efficient conditional pre-training}, which reduces the computational costs of pre-training and maintains high performance on the target task.

\section{Methods}

\begin{figure*}[!ht]
\centering
\includegraphics[width=0.72\textwidth]{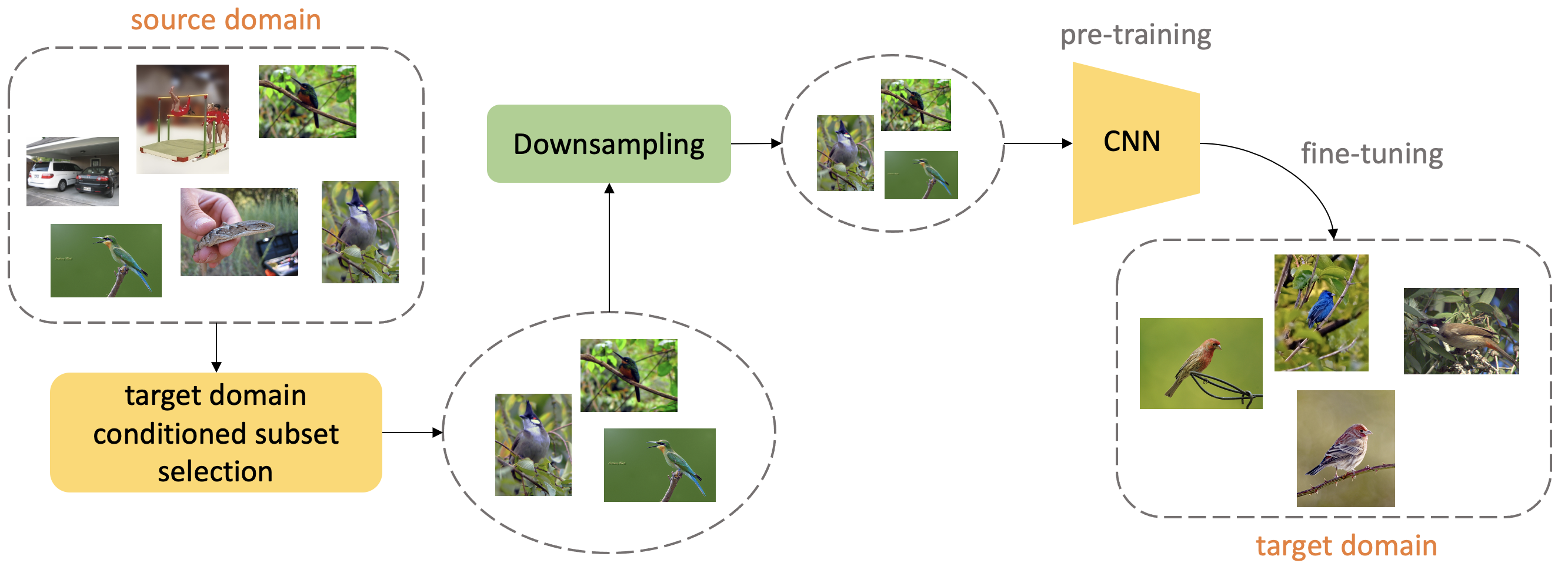}
\caption{We first perform a conditional filtering method on the source dataset and downsample image resolution on this filtered subset. Finally, we perform pre-training on the subset and finetuning on the target task.} 
\label{fig:overview}
\end{figure*}

We investigate a variety of methods to perform efficient pre-training while maintaining high performance on the target dataset. We visualize our overall procedure in Figure \ref{fig:overview} and explain our techniques below. 
\subsection{Conditional Data Filtering}
We propose novel methods to perform conditional filtering efficiently. Our methods score every image in the source domain and select the best scoring images according to a pre-specified data budget $N'$. Our methods are fast, requiring only one forward pass through $\mathcal{D}_{s}$ to get the filtered dataset $\mathcal{D}_{s}^{'}$ and can work on both $\mathcal{D}_{s}=(\mathcal{X}_{s}, \mathcal{Y}_{s})$ and $\mathcal{D}_{s}=(\mathcal{X}_{s})$. The fact that we consider \emph{data features not labels} perfectly lends our methods to the latter, more relevant, unsupervised setting. This is in contrast to previous work such as \cite{cui2018large,ge2017borrowing,ngiam2018domain} which do not consider efficiency and are designed primarily for the supervised setting
and thus will be more difficult to apply to large scale datasets. 

\begin{algorithm}[!ht]
\caption{Clustering Based Filtering}\label{alg:clustering}
\begin{algorithmic}[1]
\Procedure{ClusterFilter}{$\mathcal{D}_{s},\mathcal{D}_{t}, N^{'}, K, AggOp$}
\State $f_{h} \gets TRAIN(\mathcal{D}_{t})$\Comment{Train Feature Extractor}
\State $\mathcal{Z}_{t}\gets \{f_{h}(x_{t}^{i})\}_{i=1}^{M}$ \Comment{Target Representations}
\State $\{\hat{z}\}_{k=1}^{K} \gets K{\text-}Means(\mathcal{Z}_{t}, K)$\Comment{Cluster Target}
\State $d_{k}^{i} \gets ||f_{h}(x_{s}^{i}) - \hat{z}_{k}||_2$\Comment{Source Distances}
\State $c_{s} \gets \{AggOp(\{d_{k}^{i}\}_{k=1}^{K})\}_{i=1}^{N}$\Comment{Score Source}
\State $\mathcal{D}_{s}^{'} \gets BOTTOM(\mathcal{D}_{s}, N^{'}, c_{s})$\Comment{Filter Source}
\State \textbf{return} $\mathcal{D}_{s}^{'}$\Comment{Return the Filtered Subset}
\EndProcedure
\end{algorithmic}
\end{algorithm}
\vspace{-1.5em}
\subsubsection{Conditional Filtering by Clustering}
Selecting an appropriate subset $\mathcal{D}_{s}^{'}$ of pre-training data $\mathcal{D}_{s}$ can be viewed as selecting a set of data that minimizes some distance metric between $\mathcal{D}_{s}^{'}$ and the target dataset $\mathcal{D}_{t}$, as explored in \cite{cui2018large, ge2017borrowing}. This is accomplished by taking feature representations $\mathcal{Z}_{s}$ of the set of images $\mathcal{X}_{s}$ and selecting pre-training image classes which are close (by some distance metric) to the representations of the target dataset classes. Building on this, we make several significant modifications to account for our goals of efficiency and application to unsupervised settings. 

\noindent \textbf{Training Only with Target Dataset.}
We do not train a network $f_{h}$ on a large scale dataset, i.e. JFT-300M \cite{cui2018large}, as this defeats the entire goal of pre-training efficiency. Therefore, we first train a model $f_{h}$ with parameters $\theta_h$ using the target dataset $\mathcal{D}_{t}=(\mathcal{X}_{t}, \mathcal{Y}_{t})$ and use the learned $\theta_h$ to filter the source dataset $\mathcal{D}_{s}$.

\noindent \textbf{Consider Source Images Individually.} Selecting entire classes of pre-training data can be suboptimal when limited to selecting a small subset of the data. For example, if limited to 6\% of ImageNet, (a reasonable budget for massive datasets), we can only select 75 of the 1000 classes, which may prohibit the model from having the breadth of data needed to learn transferable features. Instead, we treat each image $x_{s}^{i}$ from $\mathcal{D}_{s}$ separately to flexibly over-represent relevant classes while not being forced to select full set of images from different classes. Additionally, very large scale datasets may not have class labels $\mathcal{Y}_{s}$. For this reason, we develop methods that work with unsupervised learning, and treating source images independently accomplishes this.

\noindent \textbf{Scoring and Filtering.} Finally, we choose to perform K-Means clustering on the representations $\mathcal{Z}_{t}$ learned by $f_{h}$ to get $K$ cluster centers $\{\hat{z}\}_{k=1}^{K}$. We then compute the distances between $\mathcal{X}_{s}$ and $\{\hat{z}\}_{k=1}^{K}$ as 
\begin{equation}
    d_{k}^{i}(x_{s}^{i},k) = ||f_{h}(x_{s}^{i};\theta_h) - \hat{z}_{k}||_p
\end{equation}
where p is typically 1 or 2 (L1 or L2 distance). We can score $x_{s}^{i}$ by considering an \emph{Aggregation Operator(AggOp)} of either average distance to the cluster centers
\begin{equation}
    c_{s}^{i} = \dfrac{1}{K}\sum_{k=1}^{K} d_{k}^{i}
\end{equation}
or minimum distance
\begin{equation}
    c_{s}^{i} = \min (\{d_{k}^{i}\}_{k=1}^{K}).
\end{equation}
\begin{figure*}[!h]
\centering
\includegraphics[width=0.80\linewidth]{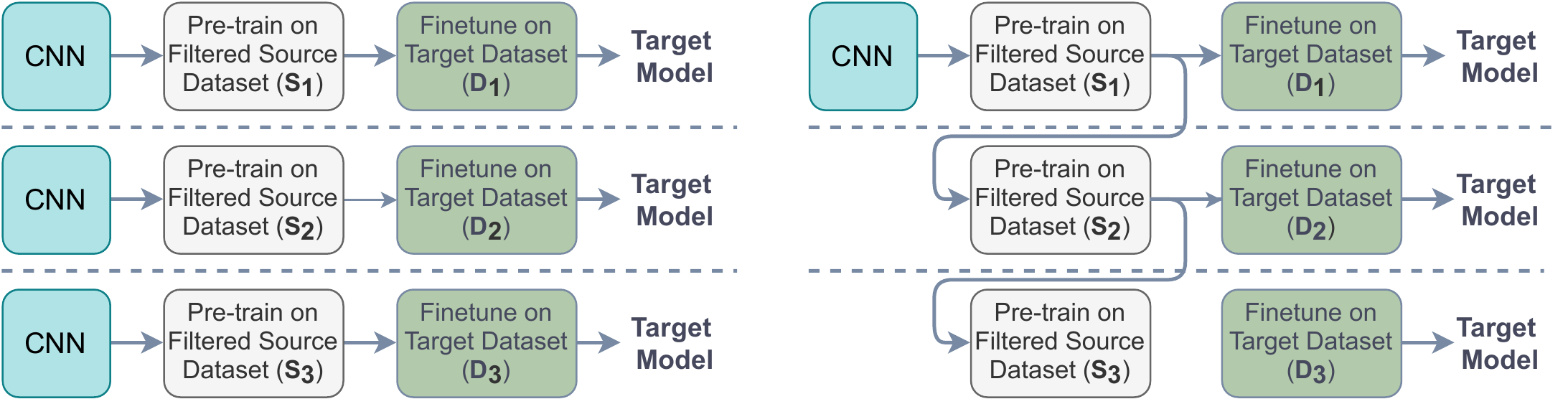}
\caption{Independent conditional pre-training (\textbf{Left}) and sequential conditional pre-training (\textbf{Right}) with $P=3$ target tasks. Sequential pre-training reduces the number of epochs required to pre-train models to accomplish these tasks.}
\label{fig:independent_vs_sequential}
\end{figure*}
To filter, we sort by $c_{s}^{i}$ in ascending order and select $N'$ images to create  $\mathcal{D}_{s}^{'} \in \mathcal{D}_{s}$ and pre-train $\theta_{t}$ on it.

\noindent \textbf{Advantages of our Method} Performing unsupervised clustering ensures that our method is not fundamentally limited to image recognition target tasks and also does not assume that source dataset images in the same class should be grouped together. Furthermore, our method requires only a single forward pass through the much smaller pre-training dataset. It attains our goals of efficiency and flexibility, in contrast to prior work such as \cite{ge2017borrowing, cui2018large}. We outline the algorithm step-by-step in Algorithm \ref{alg:clustering} and lay out the method visually in the \textbf{Appendix}.

\begin{algorithm}[!ht]
\caption{Domain Classifier Filtering}\label{alg:classifier}
\begin{algorithmic}[1]
\Procedure{DomainClsFilter}{$\mathcal{D}_{s},\mathcal{D}_{t}, N^{'}$}
\State SAMPLE $\{x_{s}^{i}\}_{i=1}^{M} \in \mathcal{D}_{s}$
\State $\mathcal{X}_h\gets \{\{x_{s}^{i}\}_{i=1}^{M},\{x_{t}^{i}\}_{i=1}^{M}\}$ 
\State $\mathcal{Y}_h\gets \{\{0\}_{i=1}^{M},\{1\}_{i=1}^{M}\}$ \Comment{Domain Labels}
\State $\mathcal{D}_{h} \gets (\mathcal{X}_{h}, \mathcal{Y}_h)$\Comment{Training Data}
\State $f_{h}(x; \theta_{h}) \gets \argmin_{\theta_{h}}CELoss(\mathcal{D}_{h})$\Comment{Fit Model}
\State $c_{s} \gets \{f_{h}(x_{s}^{i}; \theta_{h})\}_{i=1}^{N}$\Comment{Score}
\State $\mathcal{D}_{s}^{'} \gets TOP(\mathcal{D}_{s}, N^{'}, c_{s})$\Comment{Filter Source}
\State \textbf{return} $\mathcal{D}_{s}^{'}$\Comment{Return the Filtered Subset}
\EndProcedure
\end{algorithmic}
\end{algorithm}

\vspace{-1.5em}
\subsubsection{Conditional Filtering with Domain Classifier}
In this section, we propose a novel domain classifier to filter $\mathcal{D}_{s}$ with several desirable attributes. We outline the algorithm step-by-step in Algorithm \ref{alg:classifier} and provide a depiction in the \textbf{Appendix}. 

\noindent \textbf{Training.} In this method, we propose to learn $\theta_h$ to ascertain whether an image belongs to $\mathcal{D}_{s}$ or $\mathcal{D}_{t}$. $\theta_{h}$ is learned on a third dataset $\mathcal{D}_{h}=(\mathcal{X}_{h}, \mathcal{Y}_h)$ where $\mathcal{X}_h = \{\{x_{s}^{i}\}_{i=1}^{M},\{x_{t}^{i}\}_{i=1}^{M}\}, M = |\mathcal{D}_{t}|$, consisting of full set of $\mathcal{D}_{t}$ and a small random subset of $\mathcal{D}_{s}$. Each source image $x_{s}^{i} \in \mathcal{X}_{s}^{'}$ receives a negative label and each target image $x_{t}^{i} \in \mathcal{X}_{t}^{'}$ receives a positive label giving us the label set $\mathcal{Y}_h = \{\{0\}_{i=1}^{M},\{1\}_{i=1}^{M}\}$. We then learn $\theta_{h}$ on $\mathcal{D}_{h}$ using cross entropy loss as
\begin{equation}
\small
    \argmin_{\theta_{h}}\sum_{i=1}^{2M}y_{h}^{i}log(f_h(x_{h}^{i};\theta_{h})) + (1-y_{h}^{i})log(1 - f_h(x_{h}^{i};\theta_{h})).
\end{equation}

\noindent \textbf{Scoring and Filtering.} Once we learn $\theta_{h}$ we obtain the confidence score $p(y_h=1|x_s^{i};\theta_h)$ for each image $x_{s}^{i} \in \mathcal{X}_{s}$. We then sort the source images $\mathcal{X}_{s}$ in descending order based on $p(y_h=1|x_s^{i};\theta_h)$ and choose the top $N'$ images to create the subset $\mathcal{D}_{s}^{'} \in \mathcal{D}_{s}$. 

\noindent \textbf{Interpretation.} Our method can be interpreted as selecting images from the pre-training dataset with high probability of belonging to the target domain. It can be shown ~\cite{grover2019bias} that the Bayes Optimal binary classifier $\hat{f_{h}}$ assigns probability

\begin{equation}
    p(y_h=1|x_s^{i};\theta_h)  = \frac{p_{t}(x_{s}^{i})}{p_{s}(x_{s}^{i})+p_{t}(x_{s}^{i})}
\end{equation} 
for an image $x_{s}^{i} \in \mathcal{X}_{s}$ to belong to the target domain, where $p_{t}$ and $p_{s}$ are the true data probability distributions for the target and source domains respectively. 


\begin{algorithm}[!ht]
\caption{Sequential Pre-training}\label{alg:sequential_pretraining}
\begin{algorithmic}[1]
\Procedure{SequentialPre-Train}{$\mathcal{T}, \mathcal{T}_{sem}, \mathcal{S}, N^{'}$}
\State $f = \text{RAND}()$ \Comment{Randomly Initialize Model}
\While{\texttt{$True$}} \Comment{Handle All Tasks}
\State $\mathcal{T}_{sem}.\text{wait}()$ \Comment{Wait for Task Semaphore} 
\State $\mathcal S, {D}_{i} = \mathcal{T}.\text{pop}()$ \Comment{Current Task from Queue}
\State $\mathcal{S}_{i}' = \text{FILTER}(\mathcal{D}_{i}, \mathcal{S}, N^{'})$
\State $f = \text{TRAIN}(f, \mathcal{S'}_{i})$ \Comment{Update Model}
\State $\text{TASK}(f, \mathcal{D}_{t}, T_{i})$ \Comment{Perform Current Task}
\EndWhile
\EndProcedure
\end{algorithmic}
\end{algorithm}

\subsection{Sequential Pre-training}
Previously, we treated pre-training for different target tasks independently by pre-training a model from scratch on each conditionally filtered source dataset.
In practice, we may be interested in many different target tasks over time, and performing separate pre-training from scratch for each one may hinder efficiency by re-learning basic image features. To avoid it, we propose sequential pre-training where we leverage previously trained models to more quickly pre-train on the next conditionally filtered source dataset.

Formally, we assume that we have a large scale source dataset $S$ (which can potentially grow over time) and want to perform tasks on P target datasets, which we receive sequentially over time as $((S, D_{1}, t_{1}), (S, D_{2}, t_{2}), \dots, (S, D_{P}, t_{P}))$. We receive our first target task with target dataset $D_{1}$ at time $t_{1}$, and we conditionally filter $S$ into $S_{1}'$ based on our data budget. Then, we pre-train a model from scratch on $S_{1}'$ and finetune it on $D_{1}$ to get target model $f_{t_{1}}$. Generally, when we receive $D_{i}$ at time $t_{i}$, we filter $S$ conditioned on $D_{i}$ to obtain $S_{i}'$. Then, we take our last pre-training model, trained on $S_{i-1}'$, and update its weights by pre-training on $S_{i}'$ and finetune on $D_{i}$ to obtain $f_{t_{i}}$ to accomplish the current task. Subsequent tasks require smaller and smaller amounts of additional pre-training, thus drastically reducing the total number of pre-training epochs required to accomplish these tasks. We lay out this procedure step by step in Algorithm~\ref{alg:sequential_pretraining} and visual comparison between independent and sequential conditional pre-training is shown in Figure~\ref{fig:independent_vs_sequential}. 
\begin{figure*}[!h]
\centering
\begin{subfigure}[b]{0.49\textwidth}
 \centering
 \includegraphics[width=\textwidth]{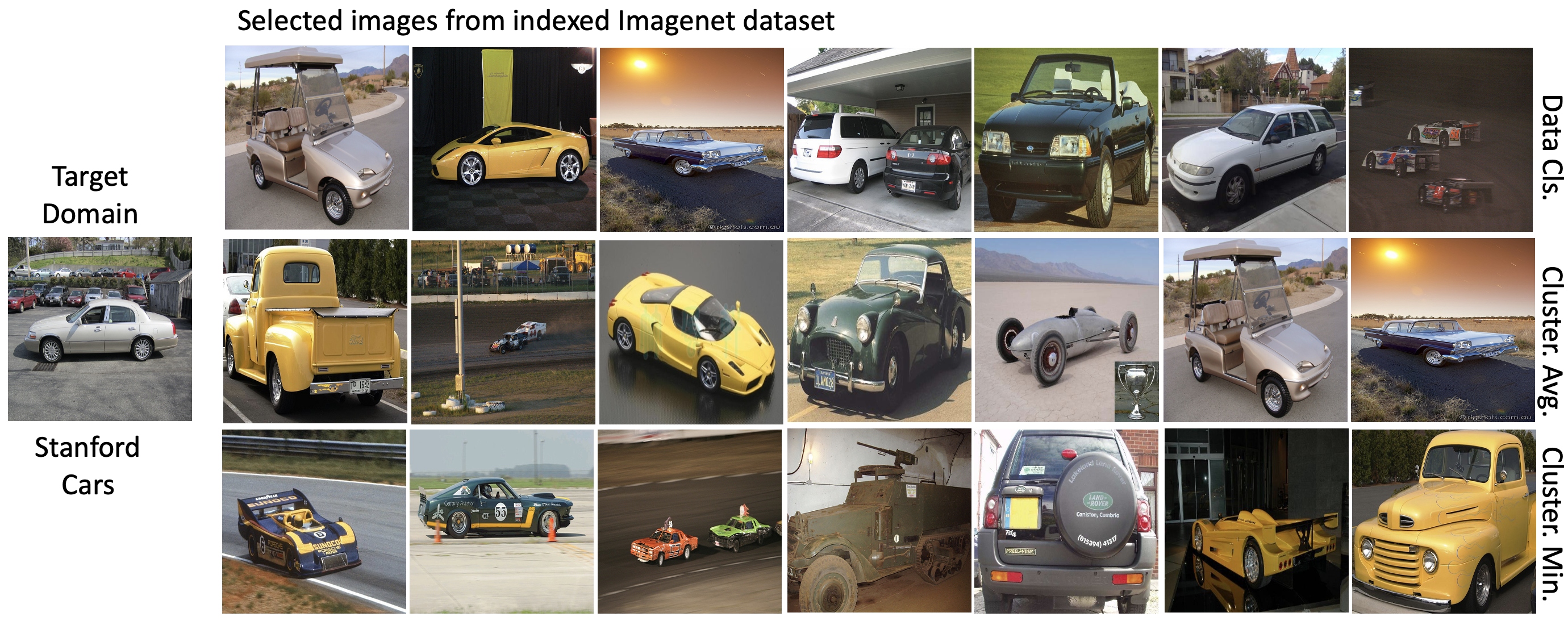}
 \label{fig:sample_cars}
\end{subfigure}
~
\begin{subfigure}[b]{0.49\textwidth}
 \centering
 \includegraphics[width=\textwidth]{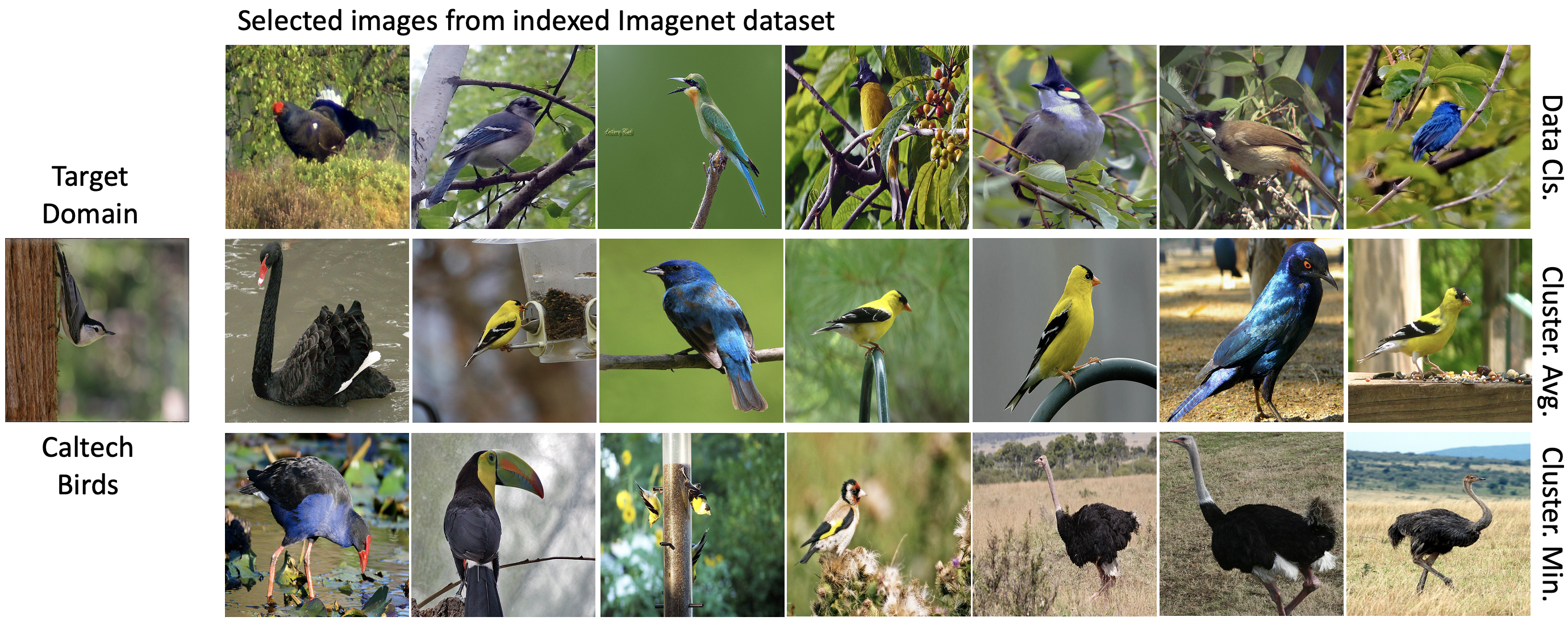}
 \label{fig:sample_birds}
\end{subfigure}
\caption{High scoring ImageNet samples selected by our conditional filtering methods for Stanford Cars and Caltech Birds.}
\label{fig:samples}
\end{figure*}
\begin{figure}
    \centering
    \includegraphics[width=\linewidth]{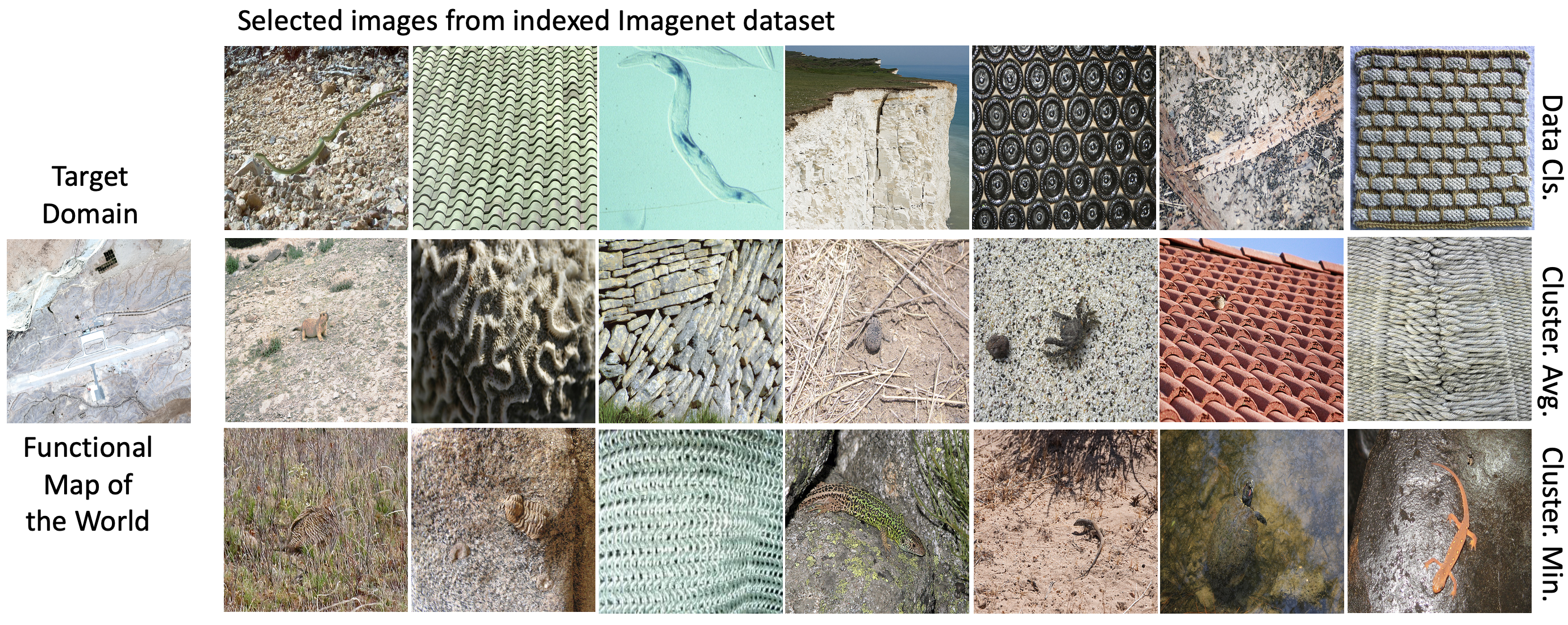}
    \caption{High scoring ImageNet samples selected by our conditional filtering methods for fMoW.}
    \label{fig:sample_fmow}
\end{figure}
\subsection{Adjusting Pre-training Spatial Resolution}
To further increase the efficiency of pre-training, we propose lowering the spatial resolution of images $\mathcal{X}_{s}$ in the source dataset $\mathcal{D}_{s}$ while pre-training. We assume that an image is represented as $x_{s}^{i} \in \mathbb{R}^{W_s\times H_s}$ or $x_{t}^{i} \in \mathbb{R}^{W_t\times H_t}$ where $W_{s}$ and $W_{t}$ represent image width in source and target dataset whereas $H_{s}$ and $H_{t}$ represent image height in source and target dataset. Traditionally, after augmentations, we use $W_{s},W_{t}=224$ and $H_{s},H_{t}=224$. Here, we consider decreasing $W_{s}$ and $H_{s}$ on the pre-training task while maintaining $W_{t},H_{t}=224$ on the target task. Reducing image resolution while pre-training can provide significant speedups by decreasing FLOPs required by convolution operations, and our experiments show that downsizing image resolution by half $W_{s},H_{s}=112$ almost halves the pre-training time with negligible loss on the target dataset.


\section{Experiments}
\subsection{Datasets}
\noindent \textbf{Source Dataset} For our primary source dataset, we utilize ImageNet-2012~\cite{deng2009imagenet}, with $\sim$1.28M images over 1000 classes. While full ImageNet is commonly used to pre-train, we use it as a proxy for a larger scale dataset that must be filtered from to thoroughly test our methods' finetuning performance under various settings. Thus, we experiment under two data budgets, limiting filtered subsets of ImageNet to 75K ($\sim$6\%) and 150K ($\sim$12\%) images. This is an appropriate proportion when dealing with pre-training datasets on the scale of tens of millions or more images. We also test our methods in larger scale settings and compile 6.71M images from the Places, OpenImages, ImageNet, and COCO datasets \cite{lin2014microsoft, zhou2017places, Kuznetsova_2020} and perform filtering on this source dataset to perform conditional pre-training.

\noindent \textbf{Image Recognition} For image recognition tasks, we utilize the Stanford Cars~\cite{WelinderEtal2010}, the Caltech Birds~\cite{krause20133d}, and a subset of the Functional Map of the World~\cite{christie2018functional} (fMoW) datasets as target datasets. We provide basic details about these datasets in the \textbf{Appendix}. These datasets lend important diversity to validate the flexibility of our methods. Cars has a fairly small distribution shift from ImageNet, and pre-training on ImageNet performs well on it, but Birds contains a larger shift and pre-training datasets emphasizing natural settings such as iNat perform better \cite{cui2018large, van2018inaturalist}. Finally, fMoW, consisting of overhead satellite images, contains images very dissimilar to ImageNet. Additionally, Birds and Cars are fine grained tasks, discriminating between different species of birds or models of cars, respectively. In contrast, fMoW contains more general categories, i.e., buildings and landmarks~\cite{uzkent2019ijcai,sarukkai2020cloud,uzkent2020learning,uzkent2020efficient}

\begin{table*}[!ht]
\begin{minipage}{0.48\hsize}
\centering
\resizebox{0.9\textwidth}{!}{%
\begin{tabular}{@{}|l|l|c|c|c|c|@{}}
\midrule
\multicolumn{2}{|c|}{\textbf{Supervised Pre-train.}} & \multicolumn{3}{c|}{\textbf{Target Dataset}} & \multicolumn{1}{c|}{\multirow{3}{*}{\textbf{\begin{tabular}[c]{@{}c@{}}Cost\\ (hrs)\end{tabular}}}} \\ \cmidrule{1-5}
\multicolumn{2}{|c|}{\textbf{224 x 224}} & \multicolumn{2}{c|}{\textbf{Small Shift}} & \multicolumn{1}{c|}{\textbf{Large Shift}} & \multicolumn{1}{l|}{} \\ \cmidrule{1-6}
\multicolumn{2}{|c|}{\textbf{Pre-train. Sel. Method}} & \multicolumn{1}{c|}{\textbf{Cars}} & \multicolumn{1}{c|}{\textbf{Birds}} & \multicolumn{1}{c|}{\textbf{fMow}} & \multicolumn{1}{l|}{} \\ \midrule
0\% & \multicolumn{1}{l|}{\textbf{Random Init.}} & \multicolumn{1}{c|}{52.89} & \multicolumn{1}{c|}{42.17} & \multicolumn{1}{c|}{43.35} & \multicolumn{1}{l|}{0} \\ \midrule
100\% & \multicolumn{1}{l|}{\textbf{Entire Dataset}} & \multicolumn{1}{c|}{82.63} & \multicolumn{1}{c|}{74.87} & \multicolumn{1}{c|}{59.05} & \multicolumn{1}{l|}{160-180} \\ \midrule
\multicolumn{1}{|c|}{\multirow{4}{*}{6\%}} & \textbf{Random} & 72.2 & 57.87 & 50.25 & 30-35 \\
\multicolumn{1}{|c|}{} & \textbf{Domain Cls.} & \textbf{74.37} & \textbf{59.73} & \textbf{51.17} & 35-40 \\
\multicolumn{1}{|c|}{} & \textbf{Clustering (Avg)} & 73.64 & 56.33 & \textbf{51.14} & 40-45 \\
\multicolumn{1}{|c|}{} & \textbf{Clustering (Min)} & \textbf{74.23} & 57.67 & 50.27 & 40-45 \\ \cmidrule{1-6}
\multirow{4}{*}{12\%} & \textbf{Random} & 76.12 & 62.73 & \textbf{53.28} & 45-50 \\
 & \textbf{Domain Cls.} & 76.18 & \textbf{64} & \textbf{53.41} & 50-55 \\
 & \textbf{Clustering (Avg)} & \textbf{77.12} & 61.73 & 53.12 &  55-60\\
 & \textbf{Clustering (Min)} & 75.81 & \textbf{64.07} & 52.91 &  55-60\\ \bottomrule
\end{tabular}%
}
\end{minipage}
\hfill
\begin{minipage}{0.48\hsize}
\centering
\resizebox{0.9\textwidth}{!}{%
\begin{tabular}{@{}|l|l|c|c|c|c|@{}}
\midrule
\multicolumn{2}{|c|}{\textbf{Supervised Pre-train.}} & \multicolumn{3}{c|}{\textbf{Target Dataset}} & \multicolumn{1}{c|}{\multirow{3}{*}{\textbf{\begin{tabular}[c]{@{}c@{}}Cost\\ (hrs)\end{tabular}}}} \\ \cmidrule{1-5}
\multicolumn{2}{|c|}{\textbf{112 x 112}} & \multicolumn{2}{c|}{\textbf{Small Shift}} & \multicolumn{1}{c|}{\textbf{Large Shift}} & \multicolumn{1}{l|}{} \\ \cmidrule{1-6}
\multicolumn{2}{|c|}{\textbf{Pre-train. Sel. Method}} & \multicolumn{1}{c|}{\textbf{Cars}} & \multicolumn{1}{c|}{\textbf{Birds}} & \multicolumn{1}{c|}{\textbf{fMow}} & \multicolumn{1}{l|}{} \\ \midrule
0\% & \multicolumn{1}{l|}{\textbf{Random Init}} & \multicolumn{1}{c|}{52.89} & \multicolumn{1}{c|}{42.17} & \multicolumn{1}{c|}{43.35} & \multicolumn{1}{l|}{0} \\ \midrule
100\% & \multicolumn{1}{l|}{\textbf{Entire Dataset}} & \multicolumn{1}{c|}{83.78} & \multicolumn{1}{c|}{73.47} & \multicolumn{1}{c|}{57.39} & \multicolumn{1}{l|}{90-110} \\ \midrule
\multicolumn{1}{|c|}{\multirow{4}{*}{6\%}} & \textbf{Random} & 72.76 & 57.4 & 49.73 & 15-20 \\
\multicolumn{1}{|c|}{} & \textbf{Domain Cls.} & 73.66 & \textbf{58.73} & 50.66 & 20-25 \\
\multicolumn{1}{|c|}{} & \textbf{Clustering (Avg)} & \textbf{74.53} & 56.97 & \textbf{51.32} & 25-30 \\
\multicolumn{1}{|c|}{} & \textbf{Clustering (Min)} & 71.72 & \textbf{58.73} & 49.06 & 25-30 \\ \cmidrule{1-6}
\multirow{4}{*}{12\%} & \textbf{Random} & 75.4 & 62.63 & 52.59 & 30-35 \\
 & \textbf{Domain Cls.} & 76.36 & \textbf{63.5} & \textbf{53.37} &  35-40 \\
 & \textbf{Clustering (Avg)} & \textbf{77.53} & 61.23 & 52.67 &  40-45 \\
 & \textbf{Clustering (Min)} & 76.36 & 63.13 & 51.6 &  40-45 \\ \bottomrule
\end{tabular}%
}
\end{minipage}
\caption{Target task accuracy and approximate filtering and pre-training cost (time in hrs on 1 GPU) on 3 visual categorization datasets obtained by pre-training on different subsets of the source dataset (ImageNet) with different filtering methods at different resolutions. \textbf{Left}: Pre-training with $224\times224$ pixels images, \textbf{Right}: Pre-training with $112\times112$ pixels images.}
\label{tab:sup_pre}
\end{table*}

\noindent \textbf{Object Detection and Image Segmentation} \cite{he2019momentum, he2020momentum} show that unsupervised ImageNet pre-training is most effective when paired with more challenging low level downstream tasks. Therefore, we also perform experiments in the object detection and semantic segmentation setting to validate the flexibility and adaptability of our methods. To this end, we utilize the Pascal VOC 2007~\cite{everingham2010pascal} dataset with unsupervised ImageNet pre-training of the backbone. 

\subsection{Analyzing Source Dataset Filtering Methods}
Here, we make some important points about our filtering methods and refer the reader to the \textbf{Appendix} for specific implementation details. 

\noindent \textbf{Domain Classifier Accuracy}
We typically train the domain classifier to $92\text{-}95 \%$ accuracy. We empirically find this is the \emph{sweet spot} as classifiers with  88-90\% accuracy, perhaps due to not learning relevant features, and 98+\% accuracy, perhaps due to over-discriminating minor differences between domains such as noise or color/contrast, do not perform as well.

\noindent \textbf{Efficiency and Adaptability Comparison.} 
The domain classifier trains a simple binary classifier and bypasses full representation learning on a target dataset, computing distances, or clustering. However, this difference in efficiency is small compared to pre-training cost. More importantly, when the target task is not image level classification, the representation learning step for clustering based filtering must be modified in a non-trivial manner. This can involve a global pooling over spatial feature maps while performing object detection or an entirely different setup like unsupervised learning. The domain classifier is more adaptable than clustering as it does not require modification for any type of target task.

\noindent \textbf{Qualitative Analysis.} In Figures \ref{fig:samples} and \ref{fig:sample_fmow}, we visualize some of the highest scoring filtered images for all our methods on image classification tasks and verify that our filtering methods do select images with \emph{relevant features} to the target task. Unsurprisingly, more interpretable images are selected for Birds and Cars, as there are no satellite images in ImageNet. Nevertheless, we see that the selected images for fMoW stil contain \emph{relevant features} such as color, texture, and shapes.

\subsection{Transfer Learning for Image Recognition}
\begin{table*}[!ht]
\begin{minipage}{0.49\hsize}
\centering
\resizebox{0.9\textwidth}{!}{%
\begin{tabular}{@{}|l|l|c|c|c|c|@{}}
\midrule
\multicolumn{2}{|c|}{\textbf{MoCo-v2~\cite{he2020momentum}}} & \multicolumn{3}{c|}{\textbf{Target Dataset}} & \multicolumn{1}{c|}{\multirow{3}{*}{\textbf{\begin{tabular}[c]{@{}c@{}}Cost\\ (hrs)\end{tabular}}}} \\ \cmidrule{1-5}
\multicolumn{2}{|c|}{\textbf{224 x 224}} & \multicolumn{2}{c|}{\textbf{Small Shift}} & \multicolumn{1}{c|}{\textbf{Large Shift}} & \multicolumn{1}{l|}{} \\ \cmidrule{1-6}
\multicolumn{2}{|c|}{\textbf{Pre-train. Sel. Method}} & \multicolumn{1}{c|}{\textbf{Cars}} & \multicolumn{1}{c|}{\textbf{Birds}} & \multicolumn{1}{c|}{\textbf{fMow}} & \multicolumn{1}{l|}{} \\ \midrule
0\% & \multicolumn{1}{l|}{\textbf{Random Init.}} & \multicolumn{1}{c|}{52.89} & \multicolumn{1}{c|}{42.17} & \multicolumn{1}{c|}{43.35} & \multicolumn{1}{l|}{0} \\ \midrule
100\% & \multicolumn{1}{l|}{\textbf{Entire Dataset}} & \multicolumn{1}{c|}{83.52} & \multicolumn{1}{c|}{67.49} & \multicolumn{1}{c|}{56.11} & \multicolumn{1}{l|}{210-220} \\ \midrule
\multicolumn{1}{|c|}{\multirow{4}{*}{6\%}} & \textbf{Random} & 75.70 & 56.82 & 52.53 & 20-25 \\
\multicolumn{1}{|c|}{} & \textbf{Domain Cls.} & 78.67 & \textbf{61.55} & 52.96 & 23-28 \\
\multicolumn{1}{|c|}{} & \textbf{Clustering (Avg)} & 78.66 & 60.88 & 53.19 & 25-30 \\
\multicolumn{1}{|c|}{} & \textbf{Clustering (Min)} & \textbf{79.45} & 59.36 & \textbf{53.5} & 25-30 \\ \cmidrule{1-6}
\multirow{4}{*}{12\%} & \textbf{Random} & 75.66 & 61.70 & 53.56 & 30-35  \\
 & \textbf{Domain Cls.} & 78.68 & 63.08 & 54.01 & 33-38 \\
 & \textbf{Clustering (Avg)} & 78.68 & 62.53 & \textbf{54.4} & 35-40 \\
 & \textbf{Clustering (Min)} & \textbf{79.55} & \textbf{63.6} & \textbf{54.26} & 35-40 \\ \bottomrule
\end{tabular}%
}
\end{minipage}
\hfill
\begin{minipage}{0.48\hsize}
\centering
\resizebox{0.9\textwidth}{!}{%
\begin{tabular}{@{}|l|l|c|c|c|c|@{}}
\toprule
\multicolumn{2}{|c|}{\textbf{MoCo-v2~\cite{he2020momentum}}} & \multicolumn{3}{c|}{\textbf{Target Dataset}} & \multicolumn{1}{c|}{\multirow{3}{*}{\textbf{\begin{tabular}[c]{@{}c@{}}Cost\\ (hrs)\end{tabular}}}} \\ \cmidrule{1-5}
\multicolumn{2}{|c|}{\textbf{112 x 112}} & \multicolumn{2}{c|}{\textbf{Small Shift}} & \multicolumn{1}{c|}{\textbf{Large Shift}} & \multicolumn{1}{l|}{} \\ \cmidrule{1-6}
\multicolumn{2}{|c|}{\textbf{Pre-train. Sel. Method}} & \multicolumn{1}{c|}{\textbf{Cars}} & \multicolumn{1}{c|}{\textbf{Birds}} & \multicolumn{1}{c|}{\textbf{fMow}} & \multicolumn{1}{l|}{} \\ \midrule
0\% & \multicolumn{1}{l|}{\textbf{Random Init}} & \multicolumn{1}{c|}{52.89} & \multicolumn{1}{c|}{42.17} & \multicolumn{1}{c|}{43.35} & \multicolumn{1}{l|}{0} \\ \midrule
100\% & \multicolumn{1}{l|}{\textbf{Entire Dataset}} & \multicolumn{1}{c|}{84.09} & \multicolumn{1}{c|}{66.57} & \multicolumn{1}{c|}{56.83} & \multicolumn{1}{l|}{110-120} \\ \midrule
\multicolumn{1}{|c|}{\multirow{4}{*}{6\%}} & \textbf{Random} & 75.38 & 56.63 & 52.59 & 10-15 \\
\multicolumn{1}{|c|}{} & \textbf{Domain Cls.} & 76.84 & 57.93 & 53.3 & 13-18 \\
\multicolumn{1}{|c|}{} & \textbf{Clustering (Avg)} & 76.86 & \textbf{58.4} & \textbf{53.75} & 15-20 \\
\multicolumn{1}{|c|}{} & \textbf{Clustering (Min)} & \textbf{77.53} & 57.1 & \textbf{53.83} & 15-20 \\ \cmidrule{1-6}
\multirow{4}{*}{12\%} & \textbf{Random} & 78.35 & 61.50 & 54.28 & 15-20 \\
 & \textbf{Domain Cls.} & \textbf{80.38} & \textbf{63.93} & 54.53 & 18-23 \\
 & \textbf{Clustering (Avg)} & 80.21 & 63.50 & \textbf{55.06} & 20-25 \\
 & \textbf{Clustering (Min)} & 79.63 & 62.77 & \textbf{55.03} & 20-25 \\ \bottomrule
\end{tabular}%
}
\end{minipage}
\caption{Target task accuracy and approximate filtering and pre-training cost (time in hrs on 4 GPUs) on 3 visual categorization datasets obtained by pre-training on different subsets of the source dataset (ImageNet) with different filtering methods at different resolutions. \textbf{Left}: Pre-training with $224\times224$ pixels images, \textbf{Right}: Pre-training with $112\times112$ pixels images.}
\label{tab:unsup_pre}
\end{table*}

\subsubsection{Supervised Pre-training Results}
We present target task accuracy for all our methods on Cars, Birds, and fMoW along with approximate pre-training and filtering time in Table \ref{tab:sup_pre}.

\noindent \textbf{Effect of Image Resolution.} We see that downsizing pre-training resolution produces gains of up to .5\% in classification accuracy on Cars and less than 1\% drop in accuracy on Birds and fMoW, while being $30\text{-}50\%$ faster than full pre-training. These trends suggest that training on lower resolution images can help the model learn more generalizeable features for similar source and target distributions. This effect erodes slightly as we move out of distribution, however pre-training on lower resolution images offers an attractive trade-off between efficiency and accuracy in all settings.

\noindent \textbf{Impact of Filtering.} We find that our filtering techniques consistently provide up to a $2.5\%$ performance increase over random selection, with a relatively small increase in cost. Unsurprisingly, filtering provides the most gains on Cars and Birds where the target dataset has a smaller shift. On fMoW, it is very hard to detect \emph{similar} images to ImageNet, as the two distributions have very little overlap. Nevertheless, in this setting, our filtering methods can still select enough relevant features to provide a $1$-$2\%$ boost.

\noindent \textbf{Comparison of Filtering Methods.} While all our methods perform well, we see that the domain classifier is less variable than clustering and always outperforms random selection. On the other hand, average clustering performs well on Cars or fMoW, but does worse than random on Birds and vice versa for min clustering. These methods rely on computing high dimensional vector distances to assign a measure of similarity, which may explain their volatility since such high dimensional distances are not considered in supervised pre-training. 
\vspace{-1em}
\subsubsection{Unsupervised Pre-training Results}
We observe promising results in the supervised setting, but a more realistic and useful setting is the unsupervised setting due to the difficulties inherent in labeling large-scale data. Thus, we use MoCo-v2~\cite{he2020momentum}, a state of the art unsupervised learning method, to pre-train on ImageNet and present results for Cars, Birds, and fMoW in Table \ref{tab:unsup_pre}.

\noindent \textbf{Effect of Image Resolution.} We find that in the unsupervised setting, with 150K pre-training images, lower resolution pre-training largely maintains or even improves performance as the target distribution shifts. Unsupervised pre-training relies more on high level features and thus may be better suited than supervised methods for lower resolution pre-training, since higher resolution images may be needed to infer fine grained label boundaries.

\noindent \textbf{Increased Consistency of Clustering.} 
Relative to the supervised setting, clustering based filtering provides more consistent boosts across the different settings and datasets. It is possible that clustering based filtering may be well suited for unsupervised contrastive learning techniques, which also rely on high dimensional feature distances. 

\noindent \textbf{Impact of Filtering.} Our filtering techniques aim to separate the image distributions based on the true image distributions and feature similarity, not label distribution (which may not be observable).
Unsupervised learning naturally takes advantage of our filtering methods, and we see gains of up to $5\%$ over random filtering in the 75K setting and up to $4\%$ in the 150K setting, a larger boost than during supervised pre-training. This leads to performance that is within $1\text{-}4\%$ of full unsupervised pre-training but close to 10 times faster, due to using a $12\%$ subset. These results are notable, because we anticipate that unsupervised learning will be the default method for large-scale pre-training and our methods can approach full pre-training while significantly reducing cost.

\vspace{-1em}
\subsubsection{Sequential Pre-training}
Cognizant of the inefficiencies of independent pre-training conditionally on the target tasks, we assume a practical scenario where over time we receive three tasks, $D_{1}, D_{2}, D_{3}$ representing Cars/Birds/fMoW respectively, with $S$ being ImageNet. We use the domain classifier to filter 150K images, obtain $S'_{1}, S'_{2}, S'_{3}$, and sequentially pre-train $f_{p}$ for 100, 40, and 20 epochs respectively with MoCo-v2. In contrast, during independent pre-training we pre-train a separate $f_{p}$ for 100 epochs for each target task.

\begin{figure}[!h]
\centering
\includegraphics[width=0.235\textwidth]{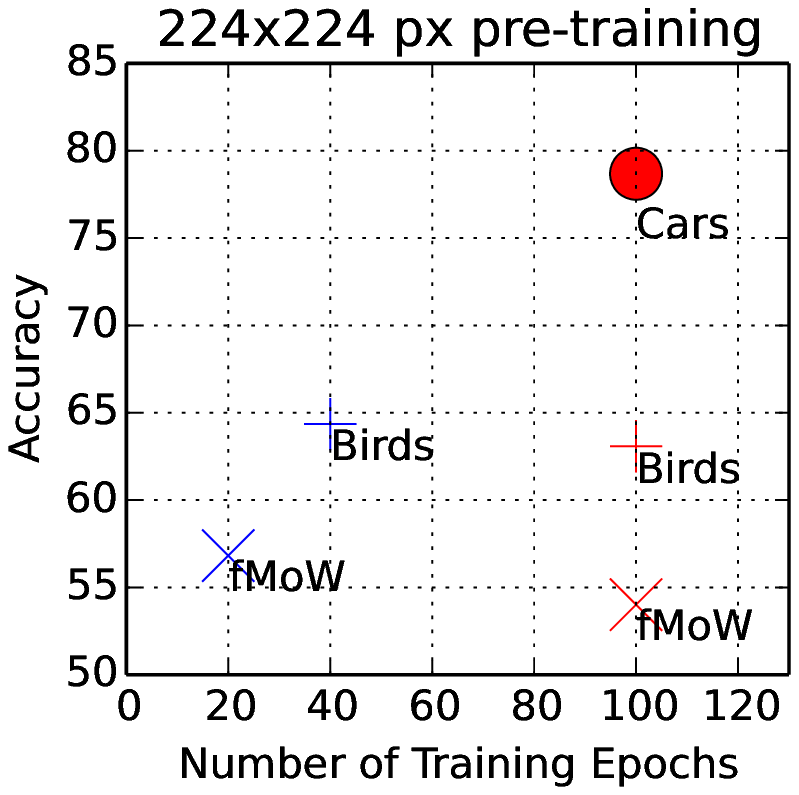}
\includegraphics[width=0.235\textwidth]{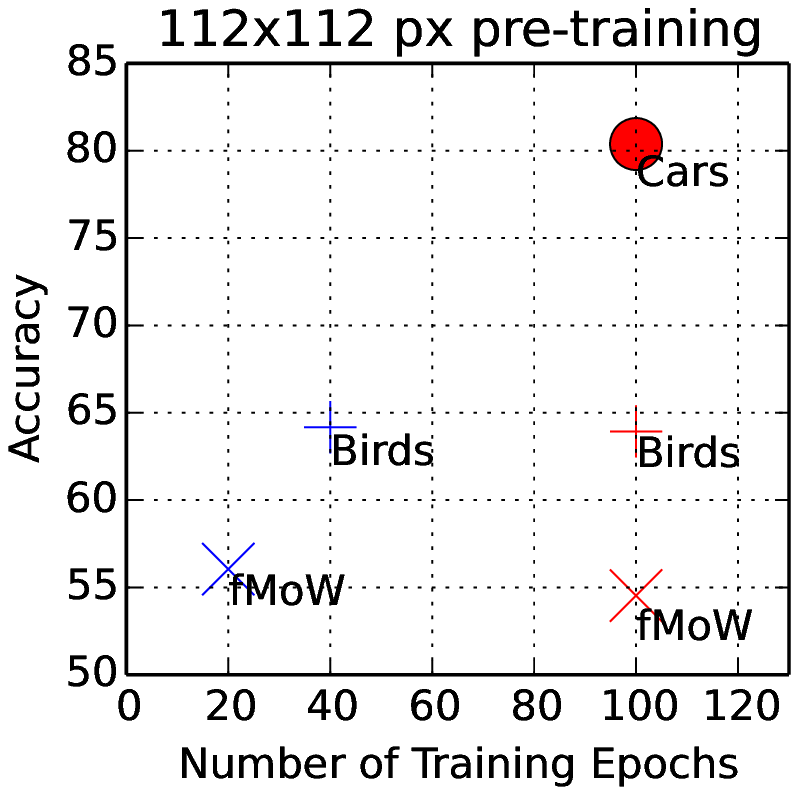}
\caption{Results for sequential pre-training (blue) vs independent pre-training (red) where pre-training with $224\times224$ pixels images is shown on the \textbf{Left} and with $112\times112$ pixels images is shown on the \textbf{Right}. Our sequential method requires fewer epochs over time with improved accuracy.} 
\label{fig:seqpretraining}
\end{figure}

We present results in Figure \ref{fig:seqpretraining}. Naturally, for Cars the results do not change, but since learned features are leveraged, not discarded, for subsequent tasks, we observe gains of up to $1\%$ on Birds and $2\%$ on fMoW over Table \ref{tab:unsup_pre} while using 160 total pre-training epochs vs 300 for independent pre-training. Our sequential pre-training method augments the effectiveness of our filtering methods in settings with many target tasks over time and drastically reduces the number of epochs required. We leave the application of this technique for object detection and segmentation as future work. 

\begin{table*}[!h]
\begin{minipage}{0.46\hsize}
\centering
\resizebox{0.98\textwidth}{!}{%
\begin{tabular}{@{}|c|l|l|l|l|l|l|l|@{}}
\toprule
\multicolumn{2}{|c|}{\textbf{Detection}} & \multicolumn{3}{c|}{\textbf{224x224}} & \multicolumn{3}{c|}{\textbf{112x112}} \\ \midrule
\multicolumn{2}{|c|}{\textbf{Pre-train. Sel. Method}} & \multicolumn{1}{c|}{\textbf{AP}} & \multicolumn{1}{c|}{\textbf{AP50}} & \multicolumn{1}{c|}{\textbf{AP75}} & \multicolumn{1}{c|}{\textbf{AP}} & \multicolumn{1}{c|}{\textbf{AP50}} & \multicolumn{1}{c|}{\textbf{AP75}} \\ \midrule
0\% & \multicolumn{1}{l|}{\textbf{Random Init.}} & \multicolumn{1}{l|}{14.51} & \multicolumn{1}{l|}{31.00} & \multicolumn{1}{l|}{11.62} & \multicolumn{1}{l|}{14.51} & \multicolumn{1}{l|}{31.00} & \multicolumn{1}{l|}{11.62} \\ \midrule
100\% & \multicolumn{1}{l|}{\textbf{Entire Dataset}} & \multicolumn{1}{l|}{43.94} & \multicolumn{1}{l|}{73.05} & \multicolumn{1}{l|}{45.96} & \multicolumn{1}{l|}{43.62} & \multicolumn{1}{l|}{72.56} & \multicolumn{1}{l|}{45.52} \\ \midrule
\multirow{4}{*}{6\%} & \textbf{Random} & 29.01 & 54.02 & 27.26 & 28.10 & 52.82 & 26.39 \\
 & \textbf{Domain Cls.} & \textbf{30.47} & \textbf{56.58} & 29.04 & \textbf{31.19} & \textbf{56.90} & \textbf{30.43} \\
 & \textbf{Clustering (Avg)} & \textbf{30.61} & 55.65 & 28.75 & 30.13 & 55.01 & 29.47 \\
 & \textbf{Clustering (Min)} & \textbf{30.44} & 56.11 & \textbf{29.46} & 30.39 & 55.89 & 28.18 \\ \cmidrule(r){1-8}
\multirow{4}{*}{12\%} & \textbf{Random} & 30.84 & 52.07 & 29.15 & 30.56 & 56.1 & 29.04 \\
 & \textbf{Domain Cls.} & \textbf{34.41} & \textbf{61.85} & \textbf{33.36} & \textbf{34.98} & \textbf{61.83} & \textbf{35.02} \\
 & \textbf{Clustering (Avg)} & 32.34 & 56.24 & 31.28 & 32.01 & 57.16 & 33.48 \\
 & \textbf{Clustering (Min)} & 32.58 & 57.77 & 31.16 & 32.96 & 58.25 & 33.64 \\ \bottomrule
\end{tabular}%
}
\end{minipage}
\hfill
\begin{minipage}{0.48\hsize}
\centering
\resizebox{\textwidth}{!}{%
\begin{tabular}{@{}|c|l|c|c|c|c|c|c|@{}}
\toprule
\multicolumn{2}{|c|}{\textbf{Segmentation}} & \multicolumn{3}{c|}{\textbf{224x224}} & \multicolumn{3}{c|}{\textbf{112x112}} \\ \midrule
\multicolumn{2}{|c|}{\textbf{Pre-train. Sel. Method}} & \multicolumn{1}{c|}{\textbf{mIOU}} & \multicolumn{1}{c|}{\textbf{mAcc}} & \multicolumn{1}{c|}{\textbf{allAcc}} & \multicolumn{1}{c|}{\textbf{mIOU}} & \multicolumn{1}{c|}{\textbf{mAcc}} & \multicolumn{1}{c|}{\textbf{allAcc}} \\ \midrule
0\% & \multicolumn{1}{l|}{\textbf{Random Init.}} & \multicolumn{1}{c|}{0.45} & \multicolumn{1}{c|}{0.55} & \multicolumn{1}{c|}{0.82} & \multicolumn{1}{c|}{0.45} & \multicolumn{1}{c|}{0.55} & \multicolumn{1}{c|}{0.82} \\ \midrule
100\% & \multicolumn{1}{l|}{\textbf{Entire Dataset}} & \multicolumn{1}{c|}{0.65} & \multicolumn{1}{c|}{0.74} & \multicolumn{1}{c|}{0.89} & \multicolumn{1}{c|}{0.63} & \multicolumn{1}{c|}{0.72} & \multicolumn{1}{c|}{0.88} \\ \midrule
\multirow{4}{*}{6\%} & \multicolumn{1}{l|}{\textbf{Random}} & \multicolumn{1}{c|}{0.55} & \multicolumn{1}{c|}{0.65} & \multicolumn{1}{c|}{0.85} & \multicolumn{1}{c|}{0.58} & 0.68 & 0.87 \\
 & \multicolumn{1}{l|}{\textbf{Domain Cls.}} & \multicolumn{1}{c|}{\textbf{0.62}} & \multicolumn{1}{c|}{\textbf{0.70}} & \multicolumn{1}{c|}{\textbf{0.88}} & \multicolumn{1}{c|}{\textbf{0.62}} & \textbf{0.70} & \textbf{0.88} \\
 & \multicolumn{1}{l|}{\textbf{Clustering (Avg)}} & \multicolumn{1}{c|}{0.61} & \multicolumn{1}{c|}{\textbf{0.70}} & \multicolumn{1}{c|}{\textbf{0.88}} & \multicolumn{1}{c|}{0.59} & 0.69 & 0.87 \\
 & \multicolumn{1}{l|}{\textbf{Clustering (Min)}} & \multicolumn{1}{c|}{0.61} & \multicolumn{1}{c|}{\textbf{0.70}} & \multicolumn{1}{c|}{\textbf{0.88}} & \multicolumn{1}{c|}{0.61} & \textbf{0.70} & \textbf{0.88} \\ \cmidrule(r){1-8}
\multirow{4}{*}{12\%} & \multicolumn{1}{l|}{\textbf{Random}} & \multicolumn{1}{c|}{0.56} & \multicolumn{1}{c|}{0.65} & \multicolumn{1}{c|}{0.86} & \multicolumn{1}{c|}{0.59} & 0.69 & 0.87 \\
 & \multicolumn{1}{l|}{\textbf{Domain Cls.}} & \multicolumn{1}{c|}{\textbf{0.65}} & \multicolumn{1}{c|}{\textbf{0.74}} & \multicolumn{1}{c|}{\textbf{0.89}} & \multicolumn{1}{c|}{\textbf{0.62}} & \textbf{0.71} & \textbf{0.89} \\
 & \multicolumn{1}{l|}{\textbf{Clustering (Avg)}} & \multicolumn{1}{c|}{0.64} & \multicolumn{1}{c|}{0.73} & \multicolumn{1}{c|}{\textbf{0.89}} & \multicolumn{1}{c|}{0.59} & 0.68 & 0.87 \\
 & \textbf{Clustering (Min)} & 0.61 & 0.70 & 0.88 & 0.61 & 0.70 & 0.88 \\ \bottomrule
\end{tabular}
}
\end{minipage}
\caption{Comparison of different source dataset filtering methods and pre-training image resolutions on transfer learning on Pascal-VOC object detection (\textbf{Left}) and semantic segmentation (\textbf{Right}) tasks. For object detection and semantic segmentation, we use unsupervised pre-training method MoCo-v2~\cite{he2020momentum}
.}
\label{tab:detection}
\end{table*}

\subsection{Transfer Learning for Low Level Tasks}
Previously we explored image level classification target tasks for conditional pre-training. In this section, we perform experiments on transfer learning for object detection and semantic segmentation on the Pascal VOC 2007 dataset. 

We present results in Table \ref{tab:detection}. For filtering, we use the domain classifier with no modifications and for clustering, we use MoCo-v2 on Pascal VOC 2007 to learn representations. We refer the reader to the \textbf{Appendix} for more experimental details. 

\noindent \textbf{Effect of Image Resolution.} Overall, pre-training on low resolution images produces no overall decrease in performance, with the corresponding $30\text{-}50\%$ reduction in training time, confirming the adaptability of pre-training on low resolution images for more challenging low level tasks.

\noindent \textbf{Adaptability Comparison} 
Relative to prior work ~\cite{cui2018large, Yan_2020_CVPR}, our clustering method is more adaptable and can efficiently be used for detection/segmentation as well as image classification. However, the representation learning step for clustering must be changed for such target tasks, which can hinder downstream performance as a representation learning technique like MoCo-v2 may be more challenging on smaller scale datasets like Pascal VOC 2007. The domain classifier, on the other hand, avoids these challenges and does not have to change when the target task is changed.

\noindent \textbf{Performance Comparison} We observe that all of our proposed filtering techniques yield consistent gains of up to 9$\%$ over random filtering, confirming their applicability to lower level tasks. In the segmentation setting, pre-training on a 12 \% subset can match full pre-training performance. Clustering produces meaningful gains, but the domain classifier outperforms it in almost every object detection scenario and the majority of segmentation metrics. This is especially pronounced with a larger pre-training subset, showing the domain classifier can filter more relevant images.

\begin{table}[!h]
\centering
\resizebox{\linewidth}{!}{%
\begin{tabular}{@{}|c|c|l|l|l|l|l|l|@{}}
\toprule
\multicolumn{2}{|c|}{\textbf{ImageNet+}} & \multicolumn{3}{c|}{\textbf{224x224}} & \multicolumn{3}{c|}{\textbf{112x112}} \\ \midrule
\multicolumn{2}{|c|}{\textbf{Pre-train. Sel. Method}} & \multicolumn{1}{c|}{\textbf{Cars}} & \multicolumn{1}{c|}{\textbf{Birds}} & \multicolumn{1}{c|}{\textbf{fMow}} & \multicolumn{1}{c|}{\textbf{Cars}} & \multicolumn{1}{c|}{\textbf{Birds}} & \multicolumn{1}{c|}{\textbf{fMow}} \\ \midrule
\multicolumn{2}{|c|}{\textbf{ImageNet}} & 83.52 & 67.49 & 56.11 & 84.09 & 66.57 & 56.83 \\ \midrule
\multicolumn{1}{|l}{} & \textbf{ImageNet+Domain Cls.} & \textbf{84.33} & \textbf{69.78} & \textbf{57.95} & \textbf{84.56} & \textbf{69.88} & \textbf{58.04} \\ \bottomrule
\end{tabular}%
}
\caption{Image classification results for ImageNet+ pre-training on three target tasks. By fine-tuning ImageNet weights on our ImageNet filtered subset, we can improve ImageNet pre-training performance on downstream classification tasks.}
\label{tab:IN+}
\end{table}

\begin{table}[]
\centering
\resizebox{\linewidth}{!}{%
\begin{tabular}{@{}|c|c|c|c|l|l|l|c|l|l|l|@{}}
\toprule
\multicolumn{1}{|l|}{} & \textbf{POIC-Random@224}       & \textbf{POIC-Ours@112} & \multicolumn{4}{c|}{\textbf{POIC-Ours@224}} & \multicolumn{4}{c|}{\textbf{ImageNet@224}} \\ \midrule
\textbf{Accuracy}      & 82.96                 & 84.29                   & \multicolumn{4}{c|}{\textbf{84.51}}                  & \multicolumn{4}{c|}{83.52}             \\ \midrule
\textbf{Cost (hrs)}       & \multicolumn{1}{c|}{210-220} & \multicolumn{1}{c|}{130-140}   & \multicolumn{4}{c|}{230-240}                       & \multicolumn{4}{c|}{210-220}                  \\ \bottomrule
\end{tabular}%
}
\caption{Results on large scale pre-training (MoCo-v2) and finetuning on the Stanford Cars dataset, with pre-training resolutions of both $112\times112$ pixels and $224\times224$ pixels. Conditionally filtering 1.28M images out of the POIC dataset with the domain classifier improves accuracy on the Stanford Cars dataset over random filtering and ImageNet (1.28M images) pre-training.
}
\label{tab:large_scale}
\end{table}
\subsection{Improving ImageNet Pre-training}
Thus far, we have used ImageNet as a proxy for a very large scale dataset to show the promise of our methods in pre-training on task-conditioned subsets. Since pre-trained models on ImageNet (1.28M images) are readily available, we now motivate practical use of our method by showing how they can outperform full ImageNet pre-training. 

\vspace{-0.12em}
\noindent \textbf{ImageNet+} Here, we take a model pre-trained on ImageNet (1.28M images) and help it focus on specific examples to our task by tuning its weights for a small number of epochs on our conditionally filtered subset of ImageNet before transfer learning. 
We apply this method to Cars/Birds/fMoW and tune pre-trained ImageNet weights with MoCo-v2 for 20 additional epochs on 150K domain classifier filtered ImageNet subsets. We present results in Table \ref{tab:IN+} and report improvements by up to $1\text{-}3\%$ over full ImageNet pre-training with minimal extra cost.

\vspace{-0.12em}
\noindent \textbf{Large Scale Filtering} Here, we envision a scenario in which a user wants to pre-train a model from scratch for a specific application with access to larger scale data than full ImageNet. To this end, we assemble a large scale dataset, which we call POIC, consisting of 6.71M images from the Places, OpenImages, ImageNet, and COCO datasets \cite{lin2014microsoft, zhou2017places, Kuznetsova_2020}. Next, we filter a subset the size of full ImageNet (1.28M images) using the domain classifier conditioned on the Cars dataset.
We pre-train the weights using an unsupervised learning method, MoCo-v2, and present our results on the Cars dataset in Table~\ref{tab:large_scale}. Our filtering methods improve on the current default of 224 resolution ImageNet pre-training by $1\text{-}1.5\%$ with good cost tradeoffs. Interestingly, a random subset of the large scale dataset performs worse than ImageNet, showing that our filtering method is crucial to select relevant examples. Here we are forced to use a 19$\%$ subset, but previous experiments showed larger relative gains for $6\%$ in comparison to $12\%$ subsets, so access to even larger scale data ($\gg$6.7M), which should be common in the future, could further improve results. This shows promise that our methods can leverage exponentially growing data scale to replace ImageNet pre-training for specific target tasks. 

\section{Conclusion}
In this work, we proposed filtering methods to efficiently pre-train on large scale datasets conditioned on transfer learning tasks including image recognition, object detection and semantic segmentation.
To further improve pre-training efficiency, we proposed decreased image resolution for pre-training and found this shortens pre-training cost by $30\text{-}50\%$ with similar transfer learning accuracy. 
Additionally, we introduced sequential pre-training to improve the efficiency of conditional pre-training with multiple target tasks. Finally, we demonstrated how our methods can improve the standard ImageNet pre-training by focusing models pre-trained on ImageNet on relevant examples and filtering an ImageNet-sized dataset from a larger scale dataset. 


{\small
\bibliographystyle{ieee_fullname}
\bibliography{egbib}

\begin{thebibliography}{10}\itemsep=-1pt

\bibitem{beluch2018power}
William~H Beluch, Tim Genewein, Andreas N{\"u}rnberger, and Jan~M K{\"o}hler.
\newblock The power of ensembles for active learning in image classification.
\newblock In {\em Proceedings of the IEEE Conference on Computer Vision and
  Pattern Recognition}, pages 9368--9377, 2018.

\bibitem{caron2020unsupervised}
Mathilde Caron, Ishan Misra, Julien Mairal, Priya Goyal, Piotr Bojanowski, and
  Armand Joulin.
\newblock Unsupervised learning of visual features by contrasting cluster
  assignments.
\newblock {\em Advances in Neural Information Processing Systems}, 33, 2020.

\bibitem{chen2020simple}
Ting Chen, Simon Kornblith, Mohammad Norouzi, and Geoffrey Hinton.
\newblock A simple framework for contrastive learning of visual
  representations.
\newblock {\em arXiv preprint arXiv:2002.05709}, 2020.

\bibitem{chen2020improved}
Xinlei Chen, Haoqi Fan, Ross Girshick, and Kaiming He.
\newblock Improved baselines with momentum contrastive learning.
\newblock {\em arXiv preprint arXiv:2003.04297}, 2020.

\bibitem{chen2020mocov2}
Xinlei Chen, Haoqi Fan, Ross Girshick, and Kaiming He.
\newblock Improved baselines with momentum contrastive learning.
\newblock {\em arXiv preprint arXiv:2003.04297}, 2020.

\bibitem{christie2018functional}
Gordon Christie, Neil Fendley, James Wilson, and Ryan Mukherjee.
\newblock Functional map of the world.
\newblock In {\em Proceedings of the IEEE Conference on Computer Vision and
  Pattern Recognition}, pages 6172--6180, 2018.

\bibitem{cui2018large}
Yin Cui, Yang Song, Chen Sun, Andrew Howard, and Serge Belongie.
\newblock Large scale fine-grained categorization and domain-specific transfer
  learning.
\newblock In {\em Proceedings of the IEEE conference on computer vision and
  pattern recognition}, pages 4109--4118, 2018.

\bibitem{deng2009imagenet}
Jia Deng, Wei Dong, Richard Socher, Li-Jia Li, Kai Li, and Li Fei-Fei.
\newblock Imagenet: A large-scale hierarchical image database.
\newblock In {\em 2009 IEEE conference on computer vision and pattern
  recognition}, pages 248--255. Ieee, 2009.

\bibitem{everingham2010pascal}
Mark Everingham, Luc Van~Gool, Christopher~KI Williams, John Winn, and Andrew
  Zisserman.
\newblock The pascal visual object classes (voc) challenge.
\newblock {\em International journal of computer vision}, 88(2):303--338, 2010.

\bibitem{gal2017deep}
Yarin Gal, Riashat Islam, and Zoubin Ghahramani.
\newblock Deep bayesian active learning with image data.
\newblock {\em arXiv preprint arXiv:1703.02910}, 2017.

\bibitem{ge2017borrowing}
Weifeng Ge and Yizhou Yu.
\newblock Borrowing treasures from the wealthy: Deep transfer learning through
  selective joint fine-tuning.
\newblock In {\em Proceedings of the IEEE conference on computer vision and
  pattern recognition}, pages 1086--1095, 2017.

\bibitem{grill2020bootstrap}
Jean-Bastien Grill, Florian Strub, Florent Altch{\'e}, Corentin Tallec, Pierre
  Richemond, Elena Buchatskaya, Carl Doersch, Bernardo Avila~Pires, Zhaohan
  Guo, Mohammad Gheshlaghi~Azar, et~al.
\newblock Bootstrap your own latent-a new approach to self-supervised learning.
\newblock {\em Advances in Neural Information Processing Systems}, 33, 2020.

\bibitem{grover2019bias}
Aditya Grover, Jiaming Song, Ashish Kapoor, Kenneth Tran, Alekh Agarwal, Eric~J
  Horvitz, and Stefano Ermon.
\newblock Bias correction of learned generative models using likelihood-free
  importance weighting.
\newblock In {\em Advances in Neural Information Processing Systems}, pages
  11058--11070, 2019.

\bibitem{he2019momentum}
Kaiming He, Haoqi Fan, Yuxin Wu, Saining Xie, and Ross Girshick.
\newblock Momentum contrast for unsupervised visual representation learning.
\newblock {\em arXiv preprint arXiv:1911.05722}, 2019.

\bibitem{he2020momentum}
Kaiming He, Haoqi Fan, Yuxin Wu, Saining Xie, and Ross Girshick.
\newblock Momentum contrast for unsupervised visual representation learning.
\newblock In {\em Proceedings of the IEEE/CVF Conference on Computer Vision and
  Pattern Recognition}, pages 9729--9738, 2020.

\bibitem{he2017mask}
Kaiming He, Georgia Gkioxari, Piotr Doll{\'a}r, and Ross Girshick.
\newblock Mask r-cnn.
\newblock In {\em Proceedings of the IEEE international conference on computer
  vision}, pages 2961--2969, 2017.

\bibitem{he2016deep}
Kaiming He, Xiangyu Zhang, Shaoqing Ren, and Jian Sun.
\newblock Deep residual learning for image recognition.
\newblock In {\em Proceedings of the IEEE conference on computer vision and
  pattern recognition}, pages 770--778, 2016.

\bibitem{hendrycks2019using}
Dan Hendrycks, Kimin Lee, and Mantas Mazeika.
\newblock Using pre-training can improve model robustness and uncertainty.
\newblock {\em arXiv preprint arXiv:1901.09960}, 2019.

\bibitem{hinton2015distilling}
Geoffrey Hinton, Oriol Vinyals, and Jeff Dean.
\newblock Distilling the knowledge in a neural network.
\newblock {\em arXiv preprint arXiv:1503.02531}, 2015.

\bibitem{huh2016makes}
Minyoung Huh, Pulkit Agrawal, and Alexei~A Efros.
\newblock What makes imagenet good for transfer learning?
\newblock {\em arXiv preprint arXiv:1608.08614}, 2016.

\bibitem{kingma2014adam}
Diederik~P Kingma and Jimmy Ba.
\newblock Adam: A method for stochastic optimization.
\newblock {\em arXiv preprint arXiv:1412.6980}, 2014.

\bibitem{kornblith2019better}
Simon Kornblith, Jonathon Shlens, and Quoc~V Le.
\newblock Do better imagenet models transfer better?
\newblock In {\em Proceedings of the IEEE conference on computer vision and
  pattern recognition}, pages 2661--2671, 2019.

\bibitem{krause20133d}
Jonathan Krause, Michael Stark, Jia Deng, and Li Fei-Fei.
\newblock 3d object representations for fine-grained categorization.
\newblock In {\em Proceedings of the IEEE international conference on computer
  vision workshops}, pages 554--561, 2013.

\bibitem{Kuznetsova_2020}
Alina Kuznetsova, Hassan Rom, Neil Alldrin, Jasper Uijlings, Ivan Krasin, Jordi
  Pont-Tuset, Shahab Kamali, Stefan Popov, Matteo Malloci, Alexander
  Kolesnikov, and et al.
\newblock The open images dataset v4.
\newblock {\em International Journal of Computer Vision}, 128(7):1956–1981,
  Mar 2020.

\bibitem{lin2014microsoft}
Tsung-Yi Lin, Michael Maire, Serge Belongie, James Hays, Pietro Perona, Deva
  Ramanan, Piotr Doll{\'a}r, and C~Lawrence Zitnick.
\newblock Microsoft coco: Common objects in context.
\newblock In {\em European conference on computer vision}, pages 740--755.
  Springer, 2014.

\bibitem{mahajan2018exploring}
Dhruv Mahajan, Ross Girshick, Vignesh Ramanathan, Kaiming He, Manohar Paluri,
  Yixuan Li, Ashwin Bharambe, and Laurens van~der Maaten.
\newblock Exploring the limits of weakly supervised pretraining.
\newblock In {\em Proceedings of the European Conference on Computer Vision
  (ECCV)}, pages 181--196, 2018.

\bibitem{ngiam2018domain}
Jiquan Ngiam, Daiyi Peng, Vijay Vasudevan, Simon Kornblith, Quoc~V Le, and
  Ruoming Pang.
\newblock Domain adaptive transfer learning with specialist models.
\newblock {\em arXiv preprint arXiv:1811.07056}, 2018.

\bibitem{park2012bayesian}
Mijung Park and Jonathan Pillow.
\newblock Bayesian active learning with localized priors for fast receptive
  field characterization.
\newblock {\em Advances in neural information processing systems},
  25:2348--2356, 2012.

\bibitem{qian1999momentum}
Ning Qian.
\newblock On the momentum term in gradient descent learning algorithms.
\newblock {\em Neural networks}, 12(1):145--151, 1999.

\bibitem{sarukkai2020cloud}
Vishnu Sarukkai, Anirudh Jain, Burak Uzkent, and Stefano Ermon.
\newblock Cloud removal from satellite images using spatiotemporal generator
  networks.
\newblock In {\em The IEEE Winter Conference on Applications of Computer
  Vision}, pages 1796--1805, 2020.

\bibitem{sener2017active}
Ozan Sener and Silvio Savarese.
\newblock Active learning for convolutional neural networks: A core-set
  approach.
\newblock {\em arXiv preprint arXiv:1708.00489}, 2017.

\bibitem{sheehan2018learning}
Evan Sheehan, Burak Uzkent, Chenlin Meng, Zhongyi Tang, Marshall Burke, David
  Lobell, and Stefano Ermon.
\newblock Learning to interpret satellite images using wikipedia.
\newblock {\em arXiv preprint arXiv:1809.10236}, 2018.

\bibitem{shin2016deep}
Hoo-Chang Shin, Holger~R Roth, Mingchen Gao, Le Lu, Ziyue Xu, Isabella Nogues,
  Jianhua Yao, Daniel Mollura, and Ronald~M Summers.
\newblock Deep convolutional neural networks for computer-aided detection: Cnn
  architectures, dataset characteristics and transfer learning.
\newblock {\em IEEE transactions on medical imaging}, 35(5):1285--1298, 2016.

\bibitem{uzkent2020learning}
Burak Uzkent and Stefano Ermon.
\newblock Learning when and where to zoom with deep reinforcement learning.
\newblock In {\em Proceedings of the IEEE/CVF Conference on Computer Vision and
  Pattern Recognition}, pages 12345--12354, 2020.

\bibitem{uzkent2018tracking}
Burak Uzkent, Aneesh Rangnekar, and Matthew~J Hoffman.
\newblock Tracking in aerial hyperspectral videos using deep kernelized
  correlation filters.
\newblock {\em IEEE Transactions on Geoscience and Remote Sensing},
  57(1):449--461, 2018.

\bibitem{uzkent2019learning}
Burak Uzkent, Evan Sheehan, Chenlin Meng, Zhongyi Tang, Marshall Burke, David
  Lobell, and Stefano Ermon.
\newblock Learning to interpret satellite images in global scale using
  wikipedia.
\newblock {\em arXiv preprint arXiv:1905.02506}, 2019.

\bibitem{uzkent2019ijcai}
Burak Uzkent, Evan Sheehan, Chenlin Meng, Zhongyi Tang, Marshall Burke, David~B
  Lobell, and Stefano Ermon.
\newblock Learning to interpret satellite images using wikipedia.
\newblock In {\em IJCAI}, pages 3620--3626, 2019.

\bibitem{uzkent2020efficient}
Burak Uzkent, Christopher Yeh, and Stefano Ermon.
\newblock Efficient object detection in large images using deep reinforcement
  learning.
\newblock In {\em The IEEE Winter Conference on Applications of Computer
  Vision}, pages 1824--1833, 2020.

\bibitem{van2018inaturalist}
Grant Van~Horn, Oisin Mac~Aodha, Yang Song, Yin Cui, Chen Sun, Alex Shepard,
  Hartwig Adam, Pietro Perona, and Serge Belongie.
\newblock The inaturalist species classification and detection dataset.
\newblock In {\em Proceedings of the IEEE conference on computer vision and
  pattern recognition}, pages 8769--8778, 2018.

\bibitem{wang2016cost}
Keze Wang, Dongyu Zhang, Ya Li, Ruimao Zhang, and Liang Lin.
\newblock Cost-effective active learning for deep image classification.
\newblock {\em IEEE Transactions on Circuits and Systems for Video Technology},
  27(12):2591--2600, 2016.

\bibitem{WelinderEtal2010}
P. Welinder, S. Branson, T. Mita, C. Wah, F. Schroff, S. Belongie, and P.
  Perona.
\newblock {Caltech-UCSD Birds 200}.
\newblock Technical Report CNS-TR-2010-001, California Institute of Technology,
  2010.

\bibitem{wu2019detectron2}
Yuxin Wu, Alexander Kirillov, Francisco Massa, Wan-Yen Lo, and Ross Girshick.
\newblock Detectron2.
\newblock \url{https://github.com/facebookresearch/detectron2}, 2019.

\bibitem{xie2015transfer}
Michael Xie, Neal Jean, Marshall Burke, David Lobell, and Stefano Ermon.
\newblock Transfer learning from deep features for remote sensing and poverty
  mapping.
\newblock {\em arXiv preprint arXiv:1510.00098}, 2015.

\bibitem{Yan_2020_CVPR}
Xi Yan, David Acuna, and Sanja Fidler.
\newblock Neural data server: A large-scale search engine for transfer learning
  data.
\newblock In {\em Proceedings of the IEEE/CVF Conference on Computer Vision and
  Pattern Recognition (CVPR)}, June 2020.

\bibitem{zhou2017places}
Bolei Zhou, Agata Lapedriza, Aditya Khosla, Aude Oliva, and Antonio Torralba.
\newblock Places: A 10 million image database for scene recognition.
\newblock {\em IEEE transactions on pattern analysis and machine intelligence},
  40(6):1452--1464, 2017.

\end{thebibliography}
}

\appendix

\section{Method Visualization}
\begin{figure*}[!ht]
\centering
\includegraphics[width=0.75\textwidth]{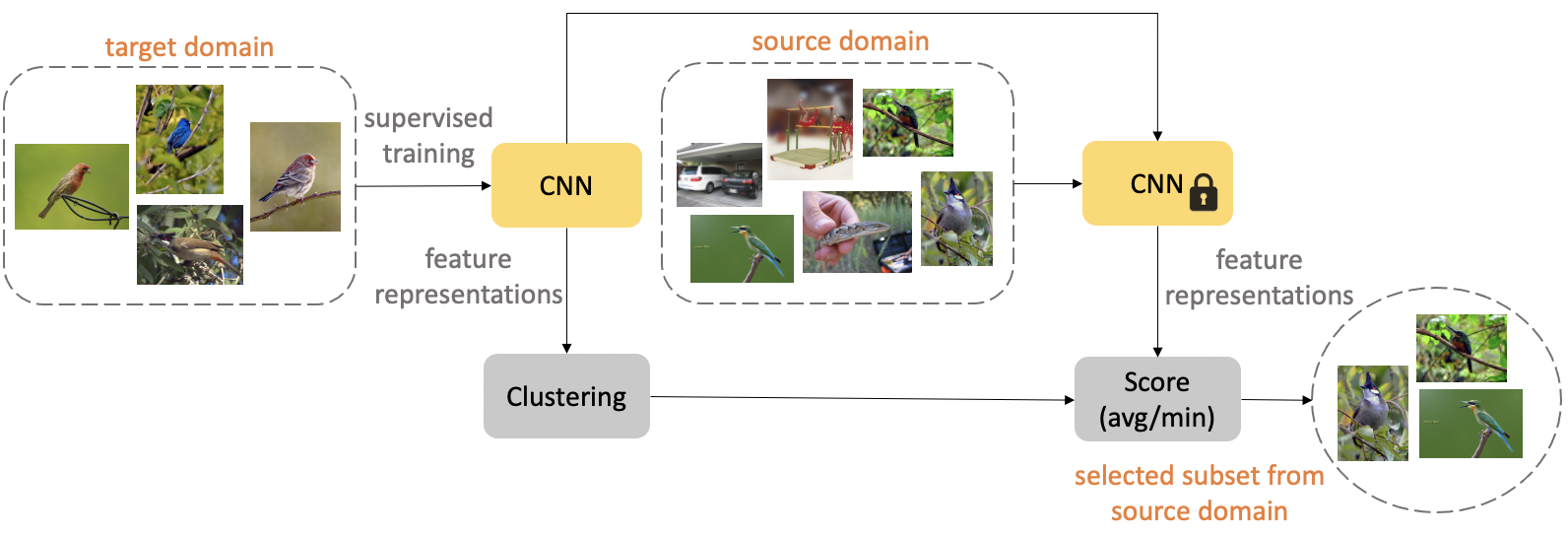}
\caption{Schematic overview of clustering based filtering. We first train a model on the target domain to extract representations, which we use to cluster the target domain. We score source images with either average or min distance to cluster centers and then filter.}
\label{fig:clustering}
\end{figure*}

\begin{figure}[!ht]
\centering
\includegraphics[width=\linewidth]{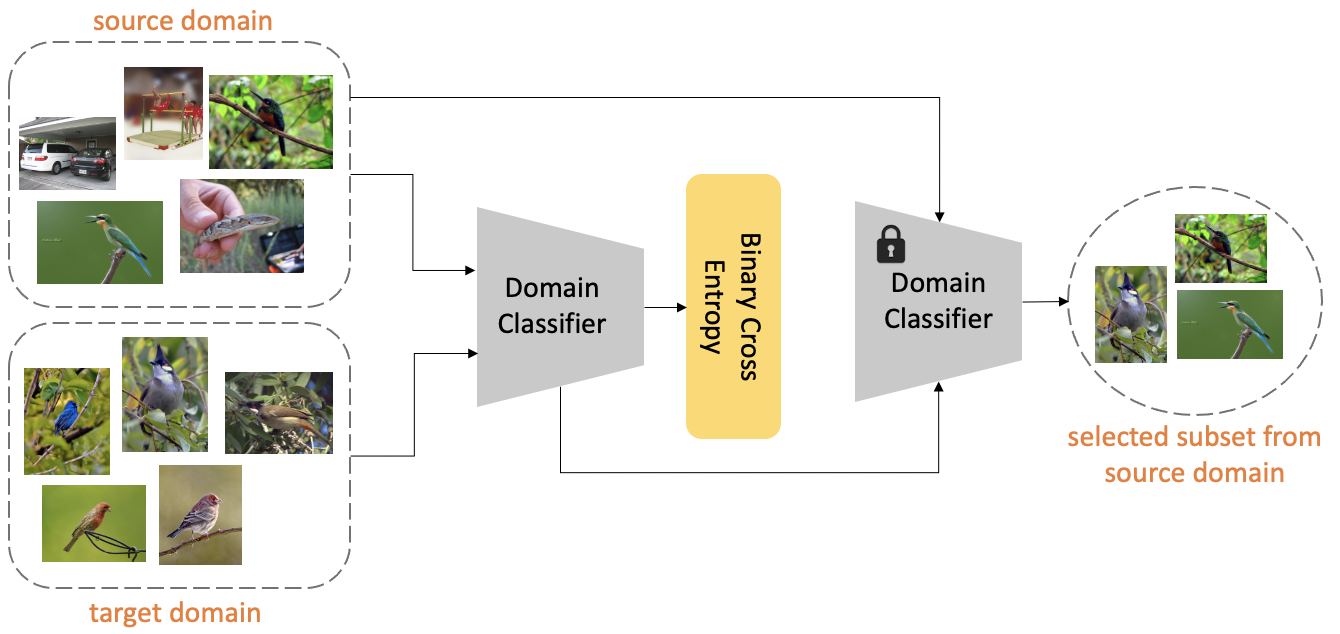}
\caption{Depiction of the Domain Classifier. We train a simple binary classifier to discriminate between source and target domain and then use the output probabilities on source images to filter.}
\label{fig:datacls}
\end{figure}

We present visual depictions for clustering based filtering in Figure \ref{fig:clustering} and for the domain classifier in Figure \ref{fig:datacls}. 

\section{Additional Methods}
\subsection{Active Learning}
Active learning is a research field concentrating on understanding which samples in a pool of samples should be given priority for annotation. One of the most common and simple active learning method relies on training a model on a labeled dataset and finding the entropy of the unseen samples by running them through the trained model. Next, top N unseen samples w.r.t their entropy (assigned by the current model) are listed in descending order. Usually, there is a single data distribution for labelled and unlabelled data, however, for our task we consider two data distributions: pre-training and target, which can be similar or completely different. For this reason, we apply two variations of active learning to conditional pre-training. First, we train a network $f_{t}$ on the target dataset $\mathcal{D}_{t}$ and run images $x_{s}^{i}$ in source dataset through the network $f_{t}$ to get the entropy of the predictions. Next, we list the images $x_{s}^{i}$ by ascending or descending entropy and choose the top $\mathcal{N}^{'}$ images. Choosing high entropy samples can be interpreted as standard active learning, and we call the method that chooses low entropy images \emph{Inverse Active Learning}.

\begin{table}[!ht]
\centering
\resizebox{\linewidth}{!}{
\begin{tabular}{|l|l|l|l|}
\midrule
\textbf{Dataset}       & \textbf{\#classes} & \textbf{\#train} & \textbf{\#test} \\ \midrule
\textbf{Stanford Cars}~\cite{WelinderEtal2010} & 196       & 8143    & 8041\\ \midrule
\textbf{Caltech Birds}~\cite{krause20133d} & 200       & 6000    & 2788 \\ \midrule
\textbf{Functional Map of the World}~\cite{christie2018functional}          & 62        & 18180   & 10609 \\ \midrule
\end{tabular}}
\caption{We use three challenging visual categorization datasets to evaluate the proposed pre-training strategies on target classification tasks.}
\label{tab:dataset}
\end{table}

\subsection{Experimental Setup}
For classification tasks, we train the linear classification layer from scratch and finetune the pre-trained backbone weights. We give basic details about our classification datasets in Table \ref{tab:dataset}. 

\textbf{Methods} We experiment with clustering based filtering, using $K=200$ clusters and both average and min distance to cluster centers, as well as our domain classifier method, using ResNet-18~\cite{he2016deep} as our classifier. Furthermore, we combine our filtering methods with downsizing pre-training image resolution from 224x224 to 112x112 using bilinear interpolation. We always perform filtering on 224 resolution source images, but use it to pre-training at both resolutions to assess flexibility, as we want robust methods that do not need to be specifically adjusted to the pre-training setup.

\textbf{Supervised Pre-training.} For supervised pre-training, in all experiments, we utilize the ResNet-34 model \cite{he2016deep} on 1 Nvidia-TITAN X GPU. We perform standard cropping/flipping transforms for ImageNet and the target data. For pre-training, we pretrain on the given subset of ImageNet for 90 epochs, utilizing SGD with momentum .9, weight decay of 1e-4, and learning rate .01 with a decay of 0.1 every 30 epochs. We finetune for 90 epochs with a learning rate decay of 0.1 every 30 epochs for all datasets. For Cars and Birds, we utilize SGD with momentum .9 \cite{qian1999momentum}, learning rate 0.1, and weight decay of 1e-4. For fMoW, we utilize the Adam optimizer \cite{kingma2014adam} with learning rate 1e-4.

\textbf{Unsupervised Pre-training.} For unsupervised pre-training, we utilize the state of the art MoCo-v2~\cite{he2020momentum} technique using a ResNet-50 model \cite{he2016deep} in all experiments. We train on 4 Nvidia GPUs. MoCo~\cite{he2019momentum,he2020momentum} is a self-supervised learning method that utilizes contrastive learning, where the goal is to maximize agreement between different views of the same image (positive pairs) and to minimize agreement between different images (negative pairs). Our choice to use MoCo is driven by (1) performance, and (2) computational cost. Compared to other self-supervised frameworks, such as SimCLR~\cite{chen2020simple}, which require a batch size of 4096, MoCo uses a momentum updated queue of previously seen samples and achieves comparable performance with a batch size of just 256 \cite{he2019momentum}. 

We keep the same data augmentations and hyperparameters used in~\cite{he2020momentum}. 
We finetune the MoCo pre-trained backbone on our target tasks for 100 epochs using a learning rate of 0.001, batch size of 64, SGD optimizer for Cars and Birds, and Adam optimizer for fMoW.

\subsection{Low Level Tasks}
\textbf{Object Detection}. We use a standard setup for object detection with a Faster R-CNN detector with a R50-C4 backbone as in \cite{he2019momentum, he2017mask, wu2019detectron2}. We pre-train the backbone with MoCo-v2 on the full or filtered subset of ImageNet. We finetune the final layers for 24k iterations ($\sim$ 23 epochs) on trainval2007 ($\sim$ 5k images). 
We evaluate on the VOC test2007 set with the default metric AP50 and the more stringent metrics of COCO-style~\cite{lin2014microsoft} AP and AP75. For filtering, we use the domain classifier with no modifications and for clustering we use MoCo-v2 on Pascal VOC to learn representations.

\textbf{Semantic Segmentation}. We use PSAnet~\cite{zhao2018psanet} network with ResNet-50 backbone to perform semantic segmentation. We train PSAnet network with a batch size of 16 and a learning rate of 0.01 for 100 epochs and use SGD optimizer. 
Similar to object detection, we pre-train the backbone with MoCo-v2 on the full or filtered subset of ImageNet and then we finetune the network using VOC train2012. We evaluate on the VOC test2012 set with the following three metrics: (a) \textbf{mIOU}: standard segmentation metric, (b) \textbf{mAcc}: mean classwise pixel accuracy, (c) \textbf{allAcc}: total pixel accuracy. For filtering, we use the domain classifier with no modifications and for clustering we use MoCo-v2 on Pascal VOC to learn representations.

\section{Additional Results}

\begin{table*}[!ht]
\begin{minipage}{0.48\hsize}
\centering
\resizebox{\textwidth}{!}{%
\begin{tabular}{@{}|l|l|c|c|c|c|@{}}
\midrule
\multicolumn{2}{|c|}{\textbf{Supervised Pre-train.}} & \multicolumn{3}{c|}{\textbf{Target Dataset}} & \multicolumn{1}{c|}{\multirow{3}{*}{\textbf{\begin{tabular}[c]{@{}c@{}}Cost\\ (hrs)\end{tabular}}}} \\ \cmidrule{1-5}
\multicolumn{2}{|c|}{\textbf{224 x 224}} & \multicolumn{2}{c|}{\textbf{Small Shift}} & \multicolumn{1}{c|}{\textbf{Large Shift}} & \multicolumn{1}{l|}{} \\ \cmidrule{1-6}
\multicolumn{2}{|c|}{\textbf{Pretrain. Sel. Method}} & \multicolumn{1}{c|}{\textbf{Cars}} & \multicolumn{1}{c|}{\textbf{Birds}} & \multicolumn{1}{c|}{\textbf{fMow}} & \multicolumn{1}{l|}{} \\ \midrule
0\% & \multicolumn{1}{l|}{\textbf{Random Init.}} & \multicolumn{1}{c|}{52.89} & \multicolumn{1}{c|}{42.17} & \multicolumn{1}{c|}{43.35} & \multicolumn{1}{l|}{0} \\ \midrule
100\% & \multicolumn{1}{l|}{\textbf{Entire Dataset}} & \multicolumn{1}{c|}{82.63} & \multicolumn{1}{c|}{74.87} & \multicolumn{1}{c|}{59.05} & \multicolumn{1}{l|}{160-180} \\ \midrule
\multicolumn{1}{|c|}{\multirow{4}{*}{6\%}} & \textbf{Random} & 72.2 & 57.87 & 50.25 & 30-35 \\
\multicolumn{1}{|c|}{} & \textbf{Inv. Active Learning} & 72.19 & 58.17 & 49.7 & 40-45 \\
\multicolumn{1}{|c|}{} & \textbf{Active Learning} & 73.17 & 57.77 & \textbf{50.91} & 40-45 \\
\multicolumn{1}{|c|}{} & \textbf{Domain Cls.} & \textbf{74.37} & \textbf{59.73} & \textbf{51.17} & 35-40 \\
\multicolumn{1}{|c|}{} & \textbf{Clustering (Avg)} & 73.64 & 56.33 & \textbf{51.14} & 40-45 \\
\multicolumn{1}{|c|}{} & \textbf{Clustering (Min)} & \textbf{74.23} & 57.67 & 50.27 & 40-45 \\
\cmidrule{1-6}
\multirow{6}{*}{12\%} & \textbf{Random} & 76.12 & 62.73 & 53.28 & 45-50 \\
 & \textbf{Inv. Active Learning} & 76.1 & 62.7 & \textbf{53.43} & 55-60 \\
 & \textbf{Active Learning} & 76.43 & 63.7 & \textbf{53.63} &  55-60\\
 &\textbf{Domain Cls.} & 76.18 & \textbf{64} & \textbf{53.41} & 50-55 \\
 & \textbf{Clustering (Avg)} & \textbf{77.12} & 61.73 & 53.12 &  55-60\\
 & \textbf{Clustering (Min)} & 75.81 & \textbf{64.07} & 52.91 &  55-60\\

  \bottomrule
\end{tabular}%
}
\end{minipage}
\hfill
\begin{minipage}{0.48\hsize}
\centering
\resizebox{\textwidth}{!}{%
\begin{tabular}{@{}|l|l|c|c|c|c|@{}}
\midrule
\multicolumn{2}{|c|}{\textbf{Supervised Pre-train.}} & \multicolumn{3}{c|}{\textbf{Target Dataset}} & \multicolumn{1}{c|}{\multirow{3}{*}{\textbf{\begin{tabular}[c]{@{}c@{}}Cost\\ (hrs)\end{tabular}}}} \\ \cmidrule{1-5}
\multicolumn{2}{|c|}{\textbf{112 x 112}} & \multicolumn{2}{c|}{\textbf{Small Shift}} & \multicolumn{1}{c|}{\textbf{Large Shift}} & \multicolumn{1}{l|}{} \\ \cmidrule{1-6}
\multicolumn{2}{|c|}{\textbf{Pretrain. Sel. Method}} & \multicolumn{1}{c|}{\textbf{Cars}} & \multicolumn{1}{c|}{\textbf{Birds}} & \multicolumn{1}{c|}{\textbf{fMow}} & \multicolumn{1}{l|}{} \\ \midrule
0\% & \multicolumn{1}{l|}{\textbf{Random Init}} & \multicolumn{1}{c|}{52.89} & \multicolumn{1}{c|}{42.17} & \multicolumn{1}{c|}{43.35} & \multicolumn{1}{l|}{0} \\ \midrule
100\% & \multicolumn{1}{l|}{\textbf{Entire Dataset}} & \multicolumn{1}{c|}{83.78} & \multicolumn{1}{c|}{73.47} & \multicolumn{1}{c|}{57.39} & \multicolumn{1}{l|}{90-110} \\ \midrule
\multicolumn{1}{|c|}{\multirow{4}{*}{6\%}} & \textbf{Random} & 72.76 & 57.4 & 49.73 & 15-20 \\
\multicolumn{1}{|c|}{} & \textbf{Inv. Active Learning} & 71.05 & 58.43 & 49.56 & 25-30 \\
\multicolumn{1}{|c|}{} & \textbf{Active Learning} & 72.95 & 56.3 & 48.94 & 25-30 \\
\multicolumn{1}{|c|}{} & \textbf{Domain Cls.} & 73.66 & \textbf{58.73} & 50.66 & 20-25 \\
\multicolumn{1}{|c|}{} & \textbf{Clustering (Avg)} & \textbf{74.53} & 56.97 & \textbf{51.32} & 25-30 \\
\multicolumn{1}{|c|}{} & \textbf{Clustering (Min)} & 71.72 & \textbf{58.73} & 49.06 & 25-30 \\
\cmidrule{1-6}
\multirow{6}{*}{12\%} & \textbf{Random} & 75.4 & 62.63 & 52.59 & 30-35 \\
 & \textbf{Inv. Active Learning} & 75.3 & 62.4 & 52.45 &  40-45 \\
 & \textbf{Active Learning} & 76.26 & 61.9 & 52.04 &  40-45 \\
 & \textbf{Domain Cls.} & 76.36 & \textbf{63.5} & \textbf{53.37} &  35-40 \\
 & \textbf{Clustering (Avg)} & \textbf{77.53} & 61.23 & 52.67 &  40-45 \\
 & \textbf{Clustering (Min)} & 76.36 & 63.13 & 51.6 &  40-45 \\
 \bottomrule
\end{tabular}%
}
\end{minipage}
\vspace{0.4em}
\caption{Results on supervised pre-training and classification tasks, including Active Learning.}
\label{tab:sup_act}
\end{table*}

\subsection{Active Learning}
We utilize our Active Learning based methods using the same supervised pre-training and finetuning setup described previously. We present our results updated with Active Learning in Table~\ref{tab:sup_act}. 

\textbf{Least vs Most Confident Samples} We see that at 224$\times$224 pixels resolution pre-training, standard active learning seems to be applicable to the transfer learning setting as selecting samples with high entropy generally does better than the inverse. 
However, at lower resolution (112$\times$112 pixels) pre-training, active learning does worse than inverse active learning and random in most cases, suggesting a lack of robustness for the active learning method since filtering is performed at 224$\times$224 pixels resolution.

\textbf{Performance Comparison} As alluded to, active learning methods perform noticeably worse in the lower resolution setting for all datasets, suggesting that filtering and pre-training conditions must be similar to maintain good performance, unlike domain classifier and clustering. In general, we see that for Cars and Birds, even at 224$\times$224 pixels resolution pre-training, active learning performance lags behind our clustering and domain classifier methods and struggles to improve over the simple random baseline in several settings. In contrast, for an out of distribution dataset like fMoW, active learning does well in the 224$\times$224 pixels resolution pre-training setting. Since active learning directly considers label distribution when filtering, it may be more prone to overfitting compared to the other methods. This can degrade its performance when relevant features are shared between the pre-training and target datasets and thus focusing only on features, not labels, when filtering may be more effective. However, in fMoW, there is very little overlap in relevant features with ImageNet, bridging the gap between active learning, and domain classifier and clustering. 

\textbf{Adaptability Comparison} Active learning is noticeably less flexible than other methods as it relies on a notion of confidence that can be hard to construct and quantify for target tasks besides classification. As said, it is also much more sensitive to filtering and pre-training resolution, making it a less robust method. In general, 
we see that 
our clustering and domain classification methods can outperform a non-trivial baseline like active learning in flexibility, adaptability, and performance.

\end{document}